\PassOptionsToPackage{table}{xcolor}
\documentclass[10pt,twocolumn,letterpaper]{article}

\usepackage[pagenumbers]{cvpr}

\usepackage{float}
\usepackage{multirow}
\usepackage{tabularx}
\newcolumntype{Y}{>{\centering\arraybackslash}X}
\usepackage{needspace}
\usepackage{wrapfig}
\usepackage{adjustbox}

\usepackage{amsmath,amsfonts,bm}

\def\eqref#1{equation~\ref{#1}}

\def\1{\bm{1}}

\DeclareMathAlphabet{\mathsfit}{\encodingdefault}{\sfdefault}{m}{sl}
\SetMathAlphabet{\mathsfit}{bold}{\encodingdefault}{\sfdefault}{bx}{n}

\graphicspath{{figures/}}


\AtBeginEnvironment{tabular}{\small}
\AtBeginEnvironment{tabular*}{\small}
\AtBeginEnvironment{tabularx}{\small}

\definecolor{groupbg}{HTML}{E8F5E9}

\definecolor{deltaup}{HTML}{2E7D32}
\definecolor{deltadown}{HTML}{C62828}
\newcommand{\dbox}[1]{\,\makebox[1.45em][l]{\fontsize{5.5pt}{6pt}\selectfont #1}}
\newcommand{\dup}[1]{\dbox{\color{deltaup}+#1}}
\newcommand{\ddown}[1]{\dbox{\color{deltadown}$-$#1}}
\newcommand{\dzero}{\dbox{\color{gray}$\pm$0.0}}
\newcommand{\dnone}{\dbox{}}

\usepackage[skins]{tcolorbox}
\definecolor{takeawayframe}{rgb}{0.21,0.49,0.74}
\newtcolorbox{takeaway}{enhanced, colback=takeawayframe!6, colframe=takeawayframe!75,
  boxrule=0.5pt, arc=1.5pt, left=4pt, right=4pt, top=4pt, bottom=5pt,
  before upper=\strut, after upper=\strut,
  before skip balanced=5pt, after skip balanced=5pt}

\newcommand{\method}{LOCUS\xspace}
\newcommand{\methodabbr}{\method}
\newcommand{\profile}{causal support profile\xspace}

\definecolor{cvprblue}{rgb}{0.21,0.49,0.74}
\usepackage[pagebackref,breaklinks,colorlinks,allcolors=cvprblue]{hyperref}

\def\confName{CVPR}
\def\confYear{2027}

\title{LOCUS: Causal Support Windows\\
for Attribute Control in\\
Autoregressive Image Generation}

\author{\normalsize\textbf{Zijun Huang}\\
\normalsize Independent Researcher}

\begin{document}
\maketitle

\setlength{\parskip}{0pt}
\begin{abstract}
Autoregressive (AR) image generators struggle to satisfy multiple attribute constraints in a single prompt.
Although these attributes are decodable from image-token latents, editing them at test time remains difficult: adding a fixed direction to the prefix can reinforce a requested attribute, yet rarely overrides a conflicting one (e.g., turning a car red despite a prompt for blue).
To find where along the image-token sequence an attribute stays controllable, we introduce the \emph{\profile}, which measures how often patching states in a given window flips the rendered attribute.
Structural attributes (count, position) are most responsive in early, narrow windows, whereas appearance attributes (color, size, material) respond over later, broader ones.
Budget-matched comparisons of early and middle windows confirm that the effective window depends on the attribute, and 2D support maps show that these regions reflect spatial structure rather than raster decode order.
Using these profiles, we propose \method (\textbf{L}atent \textbf{O}ptimization in \textbf{C}a\textbf{u}sal \textbf{S}upport windows), which diagnoses failed attributes and routes latent optimization to their support windows.
With the generator, optimizer, reward, verifier, and token budget held fixed, window placement alone creates a 23.6-point gap in GenEval++ macro accuracy between routed and late windows.
Without weight updates, \method raises Janus-Pro's GenEval++ accuracy from 24.6\% to 46.8\% overall (19.5\% to 40.5\% on multi-attribute tags).
On GenEval, it repairs 48.1\% of failures in one rollout (37.4\% for prefix optimization) in 14\% fewer rounds on average.

\noindent Code: \textup{\url{https://github.com/Huangzjun/LOCUS}}.\par

\end{abstract}
\setlength{\parskip}{0pt plus 1pt}

\section{Introduction}
\label{sec:intro}

\begin{figure}[t]
\centering
\includegraphics[width=0.98\linewidth]{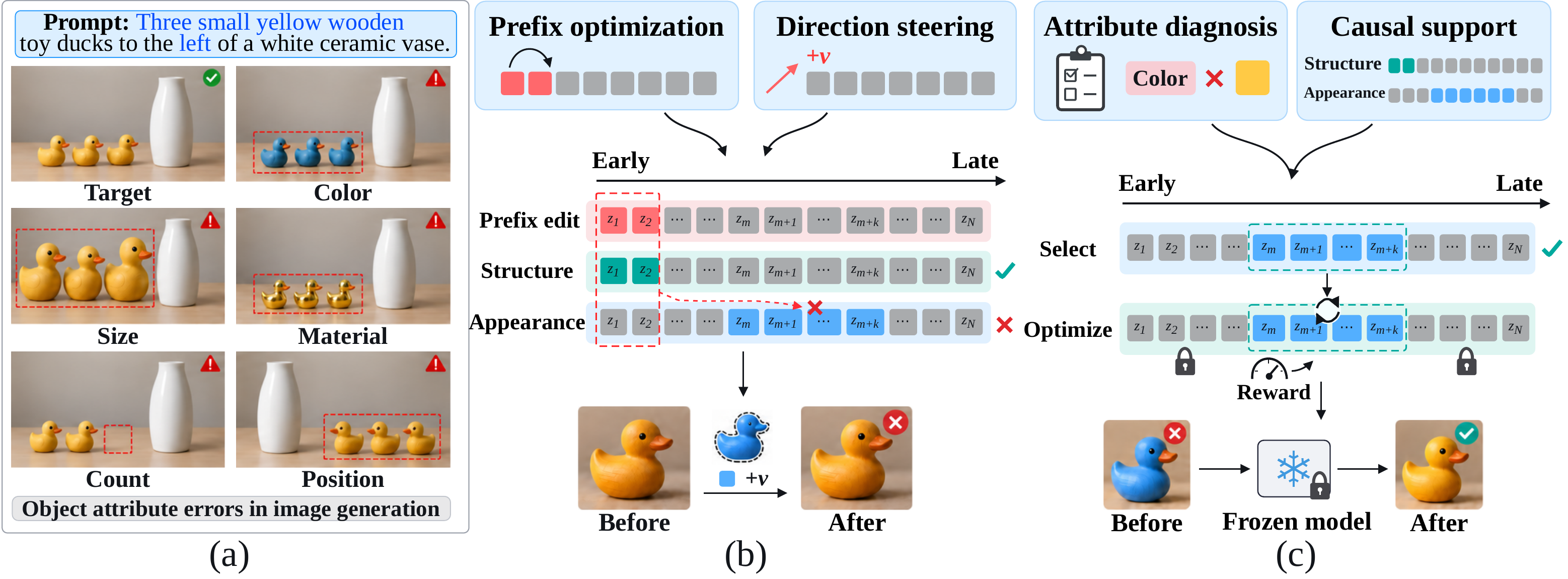}
\caption{\textbf{Motivation of \method.}
(a)~AR generators often get object attributes wrong, such as color, size, material, count, and position, especially when a prompt combines them.
(b)~Prefix-optimization and direction-steering methods edit only the earliest latents and can miss attributes whose support window lies later, such as appearance, while (c)~\method diagnoses the failed attribute and optimizes only the latents in its support window.}
\label{fig:teaser}
\end{figure}

Autoregressive (AR) image generators such as LlamaGen~\citep{sun2024autoregressive} and Emu3~\citep{wang2024emu3} emit an image as a sequence of discrete tokens, and unified models such as Janus-Pro~\citep{chen2025janus} now understand and generate images with a single AR backbone.
Yet these models still get \emph{object attributes} wrong, and most often on compositional prompts that bind several of them at once~\citep{meral2024conform,huang2025t2icompbenchpp,binyamin2025count,huang2026prove} (Figure~\ref{fig:teaser}a).
\emph{Appearance} attributes (\textit{color}, \textit{size}, \textit{material}) describe how a single object looks; \emph{structural} attributes (\textit{count}, \textit{position}) describe how objects are laid out in the scene; a compositional prompt such as ``three red apples'' mixes the two.
On GenEval++~\citep{ye2025echo4o}, Janus-Pro scores 37.5\% on the single-attribute tags (Color, Count) but 19.5\% on the five multi-attribute tags, such as Color/Count and Pos/Size.
These are the failures a test-time repair should fix without retraining.

These models do encode the attributes.
A linear probe~\citep{alain2016probes} reads an object's color from the image-token latents with 67.1\% accuracy (chance: 10\%), and a sparse autoencoder~\citep{cunningham2023sparse,bricken2023monosemanticity} recovers 148 monosemantic color features.
Steering the prefix latents along the probe direction, however, \emph{amplifies} a color the prompt already requests, yet overrides a conflicting one (e.g., a blue car edited toward red) in only 26.7\% of trials, on par with a random direction (\S\ref{sec:phenomenon}).
The attribute is readable but not freely steerable; we call this the \emph{readability--steerability gap}, consistent with evidence that decodability does not imply causal use~\citep{elazar2021amnesic,belinkov2022probing}.

Test-time control methods inherit this gap.
Latent optimization tunes image-token latents against a reward~\citep{mi2025milr,li2025seek}, and representation steering adds a fixed concept direction to the hidden states~\citep{li2023inference}; both work for some attributes and fail for others, with no account of which.
Neither treats \emph{where} along the trajectory it intervenes as a variable: each acts at one location whichever attribute failed (Figure~\ref{fig:teaser}b).
This matters most for compositional prompts, where the failed attribute may be a count in one image and a color in the next.
AR decoding commits the image one token at a time and freezes each state in the KV-cache, so each attribute may stay changeable only within its own window of the sequence.
We ask: \emph{when and where along the AR trajectory does each object attribute become causally steerable?}

To answer it, we introduce the \emph{\profile} $C_a(W)$: given two prompts that differ only in attribute $a$, the rate at which copying the image-token states in window $W$ from one run into the other flips the rendered attribute (\S\ref{sec:setup-measure}).
Patching the full trajectory at any single layer suffices to flip color, so control is indexed by decode step, not by layer.
Across attributes, the profiles fall into two regimes: structural attributes commit early and narrowly, while appearance attributes commit later, over the object body, and barely respond to prefix edits.
The two attributes of a compositional prompt are therefore decided at different points of the sequence, and an intervention fixed at one location can reach failures of one kind but not the other.
The split is consistent with the one-directional information flow of the causal mask and KV-cache, and it suggests a testable hypothesis: \emph{a test-time repair succeeds only when the intervention overlaps the failed attribute's support window}.

\method (\textbf{L}atent \textbf{O}ptimization in \textbf{C}a\textbf{u}sal \textbf{S}upport windows) turns this hypothesis into a repair method.
It diagnoses the failed attribute and routes test-time latent optimization~\citep{mi2025milr} to that attribute's support window instead of always to the prefix (Figure~\ref{fig:teaser}c).
The same profiles explain the mixed record of prefix optimization, whose window contains the support of structural attributes and misses that of appearance attributes.
With the token budget, optimizer, reward, and verifier fixed, support-routed windows score 23.6 percentage points higher than late windows on GenEval++.
The margin over prefix optimization is largest where the profiles predict it: on the multi-attribute tags that pair a structural with an appearance attribute, \method reaches 45.0\%, against 38.3\% for the prefix and 21.7\% for the late window.
In a blind sweep that ignores the profiles, repair success still rises with support overlap (\S\ref{sec:location}).
The advantage holds under a second verifier~\citep{qwen2026qwen35} (\S\ref{sec:location-matched}), the support follows spatial content rather than raster order and transfers to held-out objects (\S\ref{sec:profile}), and the structure-before-appearance order recurs on LlamaGen-XL and Lumina-mGPT 2.0.
We make three contributions:
\begin{itemize}
\item We define the \profile of each attribute and measure it by cross-prompt patching, localizing control along the generation sequence rather than network depth.
\item We show that structural attributes commit early and narrowly and appearance attributes later and broadly, that this order recurs on two more AR models, and that support overlap predicts repair success at a fixed budget.
\item We propose \method, which routes test-time latent optimization to the support window of each failed attribute. With the generator frozen, it raises Janus-Pro's GenEval++ accuracy from 24.6\% to 46.8\% and, on GenEval~\citep{ghosh2023geneval}, improves single-rollout repair success over prefix optimization by 10.7 percentage points (\S\ref{sec:location-efficiency}).
\end{itemize}

\section{Related Work}
\label{sec:related}

\begin{figure*}[t]
\centering
\includegraphics[width=\textwidth]{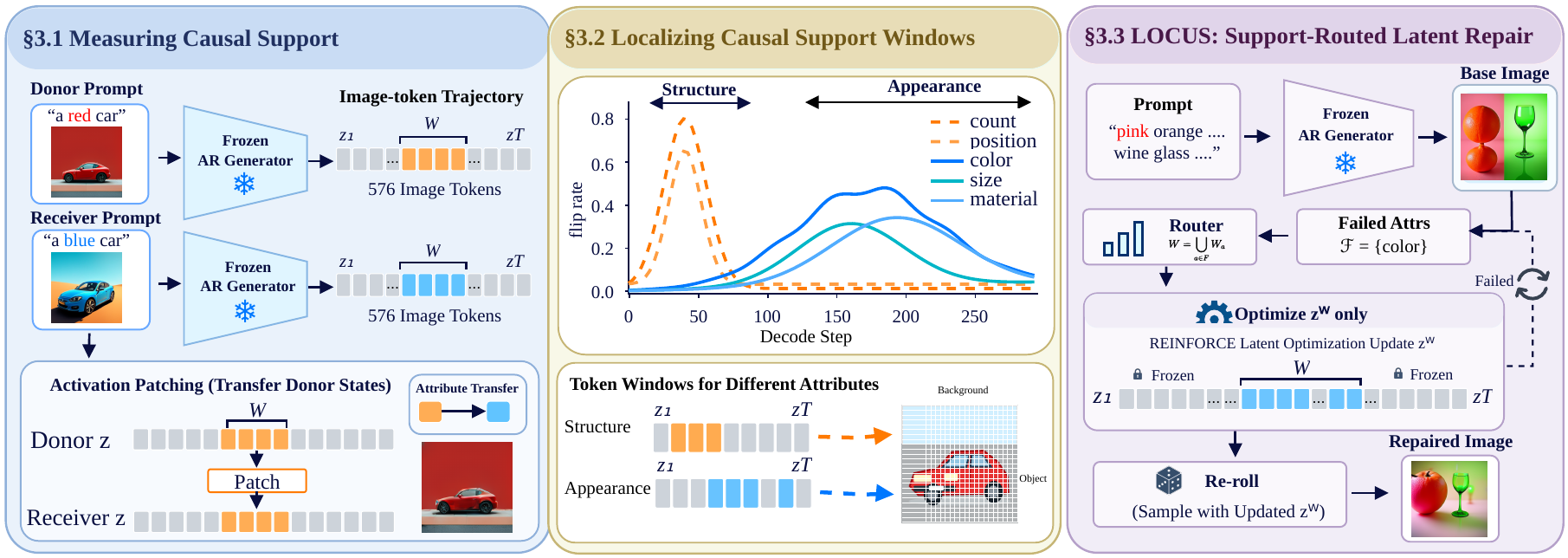}
\caption{\textbf{Overview of \method.}
\S\ref{sec:setup} measures causal support $C_a(W)$ by cross-prompt activation patching: donor hidden states replace the receiver's over a window $W$, and the flip rate is the support.
\S\ref{sec:support-profiles} locates each attribute's support window on the image-token trajectory: count and position peak early and narrowly, while color, size, and material peak in the mid band over the object body.
\S\ref{sec:location-method} diagnoses a failed generation, routes optimization to the union of the failed attributes' windows under a fixed token budget, and updates only those latents with REINFORCE while the generator stays frozen.}
\label{fig:crr}
\end{figure*}

\noindent\textbf{Autoregressive and unified image generators.}
Token-based image models cast pixels as discrete visual tokens~\citep{vandenoord2017vqvae,esser2021vqgan} and emit them autoregressively, as in LlamaGen~\citep{sun2024autoregressive}, Emu3~\citep{wang2024emu3}, Chameleon~\citep{chameleon2024}, Lumina-mGPT~\citep{liu2024luminamgpt}, and next-scale variants such as VAR~\citep{tian2024var}; unified models extend the same backbone to image understanding~\citep{chen2025janus,xie2024showo,xie2025showo2,deng2025bagel,wu2025harmon,wu2025qwenimage}.
Each image token is conditioned on the prompt and on all previous tokens, so an attribute could in principle be decided anywhere along the sequence; where it is decided has not been measured.

\noindent\textbf{Test-time control of AR image generators.}
MILR~\citep{mi2025milr} adapts LatentSeek~\citep{li2025seek} to text-to-image generation and optimizes prefix image-token latents by policy gradient against a reward.
TwiG~\citep{guo2025twig} inserts textual thoughts at predefined decoding spans, LatentMorph~\citep{chen2026latentmorph} learns when to trigger latent reasoning, ScalingAR~\citep{chen2025scalingar} schedules guidance with visual-token entropy, and other schemes prune or anchor partial decodes~\citep{park2025gridar} or plan and refine around generation~\citep{fang2025got}.
These methods use fixed or learned schedules, or treat the trajectory as one block; none identifies which image-token states are causally responsible for a given attribute.
\method keeps an existing optimizer, adds no schedule, and places the optimization inside the measured support of the failed attribute.
LAC~\citep{zhai2026lac} observes that understanding need not imply controllable generation; our measurements make it testable: control fails when the intervention misses the support window.

\noindent\textbf{Temporal structure in diffusion generation.}
Diffusion control often acts at selected denoising timesteps~\citep{meng2021sdedit,hertz2022prompt,chefer2023attendandexcite}, and theory formalizes \emph{critical windows} in which specific features emerge~\citep{li2024critical}.
Sparse-autoencoder studies find that scene composition is fixed in early denoising steps while style and texture remain editable later~\citep{tinaz2025emergence,huang2025tide}.
Our structure-early, appearance-late profile mirrors this pattern on a different axis: AR image tokens follow a spatial order, are carried through a KV-cache, and re-read the prompt at every step, so the support we study is a region of the token trajectory, not of a noise schedule.

\noindent\textbf{Causal interpretability and representation steering.}
Activation patching and causal mediation localize the components behind a behavior by transplanting activations between runs~\citep{meng2022locating,zhang2023activation,vig2020causal,wang2022ioi}.
Probes~\citep{alain2016probes} and sparse autoencoders~\citep{cunningham2023sparse,bricken2023monosemanticity} show that concepts can be linearly readable, but probe accuracy can be spurious~\citep{hewitt2019control} and decodable information can be removed without changing behavior~\citep{elazar2021amnesic}, so surveys advocate intervention-based claims~\citep{belinkov2022probing}.
Representation engineering adds fixed concept directions to hidden states~\citep{li2023inference,turner2023actadd,rimsky2024caa,zou2023repe}, and sparse-feature steering has carried this to image generators~\citep{kim2025concept}.
We bring this causal stance to the AR image-token trajectory and ask \emph{when and where} an encoded attribute becomes usable for control; the answer accounts for the readability--steerability gap, in which a readable direction amplifies an attribute yet fails to override a conflicting prompt.

\section{Method: From Causal Support to \method}
\label{sec:method}

\method is built on a measurement of where along the image-token trajectory an attribute can still be changed (Figure~\ref{fig:crr}).
We define this measurement (\S\ref{sec:setup}), use it to locate a support window for each attribute (\S\ref{sec:support-profiles}), and route test-time repair to these windows (\S\ref{sec:location-method}).

\subsection{Measuring causal support}
\label{sec:setup}
\label{sec:setup-measure}

\noindent\textbf{Model.}
We study Janus-Pro-7B~\citep{chen2025janus}, whose generator emits an image as 576 discrete visual tokens~\citep{vandenoord2017vqvae,esser2021vqgan} in raster order on a $24\times24$ grid, so step $t$ occupies row $\lfloor t/24\rfloor$.
Decoding uses classifier-free guidance~\citep{ho2022cfg}, and unless noted, interventions act on the conditional stream.
The appendix repeats the analysis on LlamaGen-XL~\citep{sun2024autoregressive} and Lumina-mGPT 2.0~\citep{liu2024luminamgpt}.

\noindent\textbf{Cross-prompt activation patching.}
We use \emph{minimal pairs} of prompts that differ in one attribute word, such as ``a \underline{red} car'' and ``a \underline{blue} car''.
We cache the per-layer residual hidden state of a \emph{donor} generation, write it into the \emph{receiver} over a layer $L$ and a step window $W$, and record whether the rendered attribute of the receiver flips to that of the donor.
For color, the pairs swap blue$\leftrightarrow$red over five objects, giving six directed pairs; the other attributes use analogous sets.
The layer sweep covers layers 0, 6, 12, 18, 24, and 29, and the window grid includes 11- and 64-step prefixes, a sliding fixed-width window, early/mid/late thirds, and the whole trajectory. Per-attribute profiles use layer 0.

\noindent\textbf{Causal support profile.}
For attribute $a$ and window $W$, the \emph{causal support profile} $C_a(W)$ is the rate at which patching $W$ flips the receiver's attribute to the donor's; we use \emph{flip rate} and \emph{override} interchangeably.
We summarize a profile by \texttt{cross05}, the number of leading tokens patched when the flip rate first reaches 50\%. A small value means the attribute commits early, and a large value means it commits late.
Because \texttt{cross05} is a cumulative-prefix summary, it is biased toward early steps, so we also report sliding fixed-width windows and a two-dimensional spatial analysis (\S\ref{sec:profile}), which do not share this bias.

\noindent\textbf{Evaluators.}
CLIP~\citep{radford2021clip} scores flips of appearance attributes, and GenEval detectors score flips of structural ones.
Because a pure-color frame trivially passes a color check, appearance flips are content-gated: a flip counts only if CLIP image--image similarity to the baseline stays within $0.92\times$.

\subsection{Localizing causal support windows}
\label{sec:support-profiles}

We first locate the readability--steerability gap at the prefix, then show that sequence coverage rather than layer limits control, and finally show that the support window depends on the attribute, is spatial, and predicts which edits succeed.
The output is a support window $W_a$ for each attribute class, which \method reads at test time.

\begin{figure}[t]
\centering
\includegraphics[width=\linewidth]{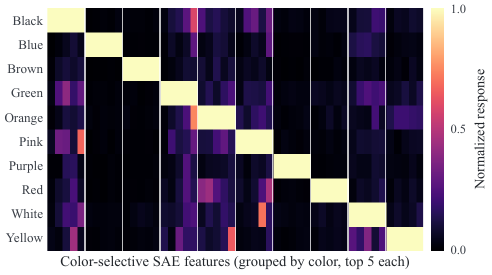}
\caption{\textbf{Color is cleanly \emph{readable} from the latents.}
SAE color-selective features are block-diagonal by preferred color, and the 148 recovered features together cover all ten colors.}
\label{fig:sae}
\end{figure}

\makeatletter
\newcommand{\subseclabel}[1]{\edef\@currentlabel{\thesubsection}\label{#1}}
\makeatother
\noindent\textbf{Readable, but not steerable at the prefix.}
\subseclabel{sec:phenomenon}
We take color as a running example (Table~\ref{tab:readctrl}).
A linear probe~\citep{alain2016probes} predicts the rendered color from image-token latents with 67.1\% accuracy.
A 256-unit sparse autoencoder (SAE) reconstructs the latents with a fraction of variance unexplained (FVU) of 0.037 and learns 148 monosemantic color features~\citep{bricken2023monosemanticity} that cover all ten colors (Figure~\ref{fig:sae}).
Yet a fixed color direction added to the prefix latents overrides a conflicting prompt in only 26.7\% of trials, on par with a random direction, whereas the same direction applied across the full trajectory overrides it in nearly all trials.
At the prefix, the direction thus \emph{amplifies} a color the prompt already requests but rarely \emph{overrides} one it contradicts.

\begin{table}[t]
\centering
\caption{\textbf{The readability--steerability gap for color (Janus-Pro).}
Prefix override matches a random-direction control.}
\label{tab:readctrl}
\begin{tabular*}{\linewidth}{@{\extracolsep{\fill}}lc}
\toprule
{\bf Measurement} &
{\bf Value} \\
\midrule
Linear probe accuracy (10-way) & 67.1\% \\
Chance (random guess) & 10.0\% \\
SAE reconstruction (FVU) & 0.037 \\
SAE-code color decodability & 66.3\% \\
Prefix fixed-direction override & 26.7\% \\
Full-trajectory fixed-direction override & $\approx 100\%$ \\
\bottomrule
\end{tabular*}
\end{table}

\begin{takeaway}
\textbf{Takeaway.}
Readability does not imply steerability: the same color direction fails at the prefix but succeeds across the trajectory, so location matters.
\end{takeaway}

\begin{figure}[t]
\centering
\includegraphics[width=\linewidth]{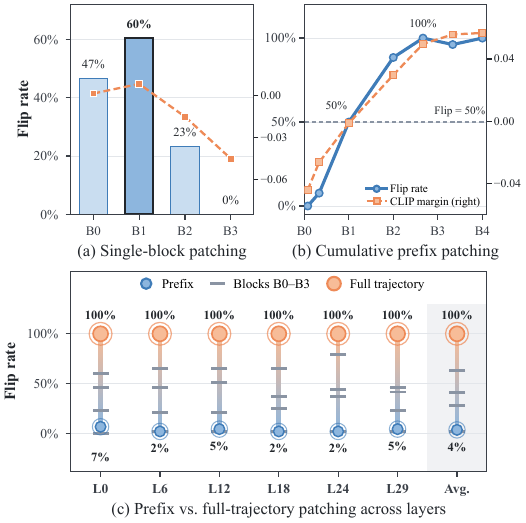}
\caption{\textbf{Color override under sequence coverage and layer choice.}
$B_k$ denotes steps $144k$--$144(k{+}1)$.
(a)~Flip rate of each block (bars) and CLIP color margin (dashed, right axis).
(b)~Flip rate and margin versus contiguous prefix coverage.
(c)~Flip rate by layer: prefix patching barely flips blue$\to$red, whereas whole-trajectory patching succeeds at every layer.}
\label{fig:coverage}
\end{figure}

\noindent\textbf{Sequence coverage, not layer, is the bottleneck.}
\subseclabel{sec:profile}
\subseclabel{sec:profile-coverage}
The prefix failure could reflect the wrong layer or too little coverage of the sequence, and a layer$\times$window sweep on color pairs separates the two (Figure~\ref{fig:coverage}).
Patching only the first 11 or 64 steps almost never flips the color at any layer, whereas patching the \emph{whole} trajectory flips it at \emph{every} single layer (Figure~\ref{fig:coverage}c).
The flip rate reaches 50\% after the first 144 steps, a quarter of the sequence (Figure~\ref{fig:coverage}b), and a sliding window shows that the block over the object body flips color most often and the last rows least often (Figure~\ref{fig:coverage}a).
A wrong-attribute donor or a wrong window flips almost nothing (appendix), so the effect is attribute-specific.

\begin{takeaway}
\textbf{Takeaway.}
Control is indexed by decode step, not by layer, and for color the prefix is the window that fails.
\end{takeaway}

\begin{figure}[t]
\centering
\includegraphics[width=\linewidth]{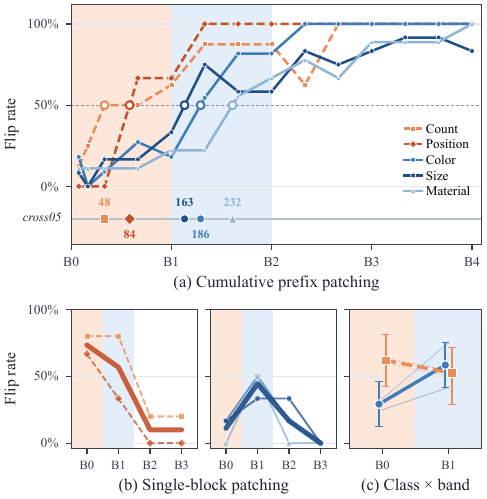}
\caption{\textbf{Commit profiles of structural (dashed) and appearance (solid) attributes.}
(a)~Flip rate versus cumulative prefix coverage; structural attributes cross 50\% at steps 48--84 and appearance attributes at 163--232.
(b)~Flip rate of each 144-step block.
(c)~Early vs.\ middle band at layer 0 (95\% bootstrap CIs).}
\label{fig:overlay}
\end{figure}

\begin{figure}[t]
\centering
\includegraphics[width=\linewidth]{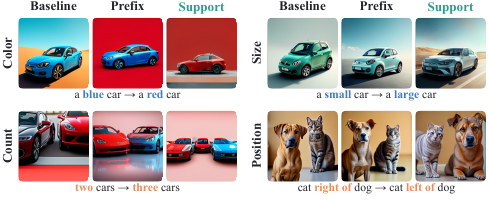}
\caption{\textbf{Qualitative patching (real generations).}
Each group shows the baseline, a prefix patch, and a support-window patch with the same donor.
Appearance (top) needs full or later coverage; structure (bottom) responds to an early window.}
\label{fig:qualpatching}
\end{figure}

\noindent\textbf{The support window depends on the attribute.}
\subseclabel{sec:profile-dichotomy}
Extending the measurement to five attributes separates two regimes (Figure~\ref{fig:overlay}a,b).
Structural attributes (\textit{count}, \textit{position}) commit early and narrowly: their flip rates reach 50\% at 48 and 84 steps, and the first 144 steps suffice to override them.
Appearance attributes (\textit{color}, \textit{size}, \textit{material}) reach 50\% only at 163--232 steps; the middle window (steps 144--288, over the object body) is the most effective, and prefix override is near zero.
A 13-attribute spectrum in the appendix places further attributes on the same early$\to$late axis.
The split is not specific to Janus-Pro: on LlamaGen-XL (256 tokens, T5 conditioning), position reaches 50\% after 6\% of the sequence (95\% CI, 16--32 tokens), whereas color and material need 25\% and 50\%, and a 16-token prefix flips position in 50\% of pairs but color in only 7\%; Lumina-mGPT 2.0 shows the same order (appendix).
Real generations show the same split (Figure~\ref{fig:qualpatching}).
A prefix color patch tints the background but leaves the object in the prompted color, and a prefix size patch leaves the object small, whereas for count and position the early window already overrides the attribute.

\begin{figure}[t]
\centering
\includegraphics[width=0.98\linewidth]{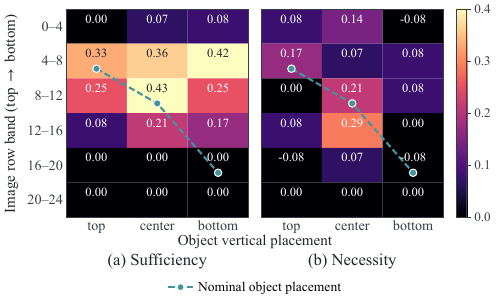}
\caption{\textbf{Color's causal support is a spatial region, not a decode-time band.}
Sufficiency and necessity over object placement $\times$ row band; bands run top$\to$bottom in decode order.}
\label{fig:supportmap}
\end{figure}

\noindent\textbf{Support is spatial, not a raster artifact.}
Because Janus-Pro decodes in raster order, ``color commits at step $K$'' could merely restate ``the object's pixels decode at step $K$''; this reading predicts support on the last-decoded bottom rows.
We place the same object at the top, center, or bottom of the image and patch one four-row band at a time over eight objects (Figure~\ref{fig:supportmap}).
Single-band patches flip color in 37\% of pairs at rows 4--8 and 32\% at rows 8--12, but never at rows 16--24, and the necessity peak shifts downward as the object moves from top to center: the support follows the object.

\noindent\textbf{The support window is predictive.}
\subseclabel{sec:profile-knockout}
\subseclabel{sec:profile-interaction}
Patching shows that a window is \emph{sufficient}, but under a generic-coverage account, in which any large enough window overrides any attribute, window location would carry no information.
A pre-specified $2\times2$ design tests this account (Figure~\ref{fig:overlay}c).
We cross attribute class (structure: \textit{position}, \textit{count}; appearance: \textit{color}, \textit{size}) with two budget-matched 144-step bands at layer 0: the early band over the image top (steps 0--144) and the middle band over the object body (steps 144--288).
If location matters, moving from the early to the middle band should help appearance but not structure.
The move raises the appearance flip rate from 29\% to 58\% and lowers the structure flip rate from 62\% to 52\%, an interaction of 39 points (95\% CI, 1--78 points; one-sided bootstrap $p=0.02$ over 5,000 resamples with 10--12 clean pairs per attribute).
The window also transfers beyond the pairs used to calibrate it: applied to 38 held-out objects with held-out colors, the color window flips 34.2\% of them, whereas an equal-length late window flips none ($p=0.0002$; appendix).

\begin{takeaway}
\textbf{Takeaway.}
No single window serves every attribute.
The prefix covers structural support but misses appearance support, and the mid band does the reverse.
\end{takeaway}

\subsection{\method: support-routed latent repair}
\label{sec:location-method}

\noindent\textbf{From findings to design.}
Because control depends on which steps an edit covers and not on the layer, \method keeps the intervention target of test-time latent optimization, the image-token latents~\citep{mi2025milr}, and varies only \emph{which steps} it optimizes.
Because no single window serves every attribute, it first identifies \emph{which} attribute failed and then intervenes on \emph{that} attribute's window.
\subseclabel{sec:profile-mechanism}
The window \emph{shape} follows an account of how information moves through the generator; the account is consistent with the profiles, although we do not measure attention directly.
The causal mask and KV-cache make the flow one-directional: a state written early is read by every later step, and a later state cannot revise it.
A spatial layout is therefore fixed once in the early states, and a sparse early window would leave gaps in which the rollout re-asserts the old layout, so position failures use a contiguous early block.
Appearance is instead re-read from the attribute word as each object region is decoded, so a budget subsampled evenly across the appearance window overlaps more of these reading steps than a contiguous block of the same size.

\noindent\textbf{Repair loop.}
Given a prompt, a frozen AR generator, a verifier, and the profiles $\{C_a\}$, \method repairs a failed base generation in rounds of four steps (Figure~\ref{fig:crr}).
(1)~\emph{Diagnose}: the verifier returns the set $F$ of failed attribute dimensions, such as count or color.
(2)~\emph{Route}: each $a\in F$ is mapped to its window $W_a$ read from $C_a$, and the union $\bigcup_{a\in F}W_a$ is subsampled evenly to a fixed token budget. Position failures use a contiguous early block, and a failure with no flagged dimension falls back to the prefix.
(3)~\emph{Optimize}: REINFORCE latent optimization~\citep{mi2025milr} updates only the latents inside the routed window, with the prefix and suffix frozen.
(4)~\emph{Re-roll}: the generator decodes a new image from the optimized latents, and the verifier checks it again.
The loop stops when the verifier accepts the image or after $N$ rounds.
Prefix optimization acts on the same leading tokens whichever attribute failed, so it helps only when the support of that attribute overlaps the prefix, as for structure but not for appearance.
\method instead spends the budget on tokens that can still change the failed attribute (implementation details are in the appendix).

\begin{table*}[t]
\centering
\caption{\textbf{Accuracy on GenEval++ (\%).}
External scores are as officially reported by \citet{ye2025echo4o,zhang2025omniverifier,jiang2026genagent,kim2026fire}.
Best scores are \textbf{bold}; second-best are \underline{underlined}.}
\label{tab:mainresult}
\setlength{\tabcolsep}{6.16pt}
\begin{tabular}{lcccccccc}
\toprule
{\bf Method} &
{\bf Color} &
{\bf Count} &
{\bf Color/Count} &
{\bf Color/Pos} &
{\bf Pos/Count} &
{\bf Pos/Size} &
{\bf Multi-Count} &
{\bf Overall} \\
\midrule
SD3-Medium~\citep{esser2024sd3} & 55.0 & 50.0 & 12.5 & 35.0 & 17.5 & 15.0 & 22.5 & 29.6 \\
FLUX.1-dev~\citep{flux2024} & 35.0 & 62.5 & 15.0 & 27.5 & 20.0 & 37.5 & 22.5 & 31.4 \\
BAGEL~\citep{deng2025bagel} & 32.5 & 60.0 & 25.0 & 32.5 & 25.0 & 47.5 & 37.5 & 37.1 \\
Qwen-Image~\citep{wu2025qwenimage} & 70.0 & \textbf{90.0} & \textbf{80.0} & 52.5 & 50.0 & 70.0 & 60.0 & 67.5 \\
GPT-4o~\citep{openai2025gpt4oimage} & \textbf{90.0} & 67.5 & \underline{72.5} & 62.5 & \underline{60.0} & \textbf{80.0} & \textbf{85.0} & \textbf{73.9} \\
Echo-4o~\citep{ye2025echo4o} & \underline{80.0} & 57.5 & 55.0 & \underline{77.5} & \textbf{62.5} & \textbf{80.0} & 62.5 & 67.9 \\
ReflectionFlow~\citep{zhuo2025reflectionflow} & 40.0 & 62.5 & 27.5 & 27.5 & 20.0 & 42.5 & 32.5 & 36.1 \\
GenAgent~\citep{jiang2026genagent} & 77.5 & \underline{77.5} & 65.0 & \textbf{80.0} & \underline{60.0} & \underline{72.5} & \underline{75.0} & \underline{72.5} \\
T2I-R1~\citep{jiang2025t2ir1} & 67.5 & 32.5 & 20.0 & 35.0 & 7.5 & 25.0 & 30.0 & 31.1 \\
FiRe-GRPO~\citep{kim2026fire} & 78.0 & 49.0 & 45.0 & 57.0 & 28.0 & 36.0 & 49.0 & 49.0 \\
\midrule
Janus-Pro~\citep{chen2025janus} & 45.0 & 30.0 & 12.5 & 30.0 & 7.5 & 35.0 & 12.5 & 24.6 \\
\rowcolor{groupbg}
\textbf{Janus-Pro + \methodabbr{} (ours)} & 57.5 & 67.5 & 55.0 & 40.0 & 25.0 & 40.0 & 42.5 & 46.8 \\
\bottomrule
\end{tabular}
\end{table*}

\section{Experiments}
\label{sec:location}
\label{sec:experiments}

\begin{table*}[t]
\centering
\caption{\textbf{GenEval++ accuracy (\%) by intervention window.}
All repair arms optimize 16 token positions and share the optimizer, reward, and verifier. Per-tag scores use GPT-4.1; the last column re-scores the same images with Qwen3.5-9B. Small numbers give the change from the Janus-Pro base.}
\label{tab:matched}
\setlength{\tabcolsep}{5.24pt}
\begin{tabular}{lccccccccc}
\toprule
\multirow{2}{*}{\bf Arm} &
\multirow{2}{*}{\bf Color} & \multirow{2}{*}{\bf Count} & \multirow{2}{*}{\bf Color/Count} & \multirow{2}{*}{\bf Color/Pos} &
\multirow{2}{*}{\bf Pos/Count} & \multirow{2}{*}{\bf Pos/Size} & \multirow{2}{*}{\bf Multi-Count} &
\multicolumn{2}{c}{\bf Macro} \\
\cmidrule(lr){9-10}
& & & & & & & & {\bf GPT-4.1} & {\bf Qwen3.5} \\
\midrule
Janus-Pro (base) & 45.0\dnone & 30.0\dnone & 12.5\dnone & 30.0\dnone & 7.5\dnone & 35.0\dnone & 12.5\dnone & 24.6\dnone & 30.5\dnone \\
\midrule
Prefix & 47.5\dup{2.5} & 57.5\dup{27.5} & 47.5\dup{35.0} & 35.0\dup{5.0} & 30.0\dup{22.5} & 32.5\ddown{2.5} & 55.0\dup{42.5} & 43.6\dup{19.0} & 54.1\dup{23.6} \\
Late & 40.0\ddown{5.0} & 25.0\ddown{5.0} & 15.0\dup{2.5} & 25.0\ddown{5.0} & 7.5\dzero & 25.0\ddown{10.0} & 25.0\dup{12.5} & 23.2\ddown{1.4} & 33.8\dup{3.3} \\
Mid & 47.5\dup{2.5} & 22.5\ddown{7.5} & 22.5\dup{10.0} & 30.0\dzero & 15.0\dup{7.5} & 30.0\ddown{5.0} & 37.5\dup{25.0} & 29.3\dup{4.7} & 37.3\dup{6.8} \\
Random & 62.5\dup{17.5} & 47.5\dup{17.5} & 45.0\dup{32.5} & 32.5\dup{2.5} & 12.5\dup{5.0} & 27.5\ddown{7.5} & 57.5\dup{45.0} & 40.7\dup{16.1} & 50.9\dup{20.4} \\
\rowcolor{groupbg}
\textbf{\method (ours)} & 57.5\dup{12.5} & 67.5\dup{37.5} & 55.0\dup{42.5} & 40.0\dup{10.0} & 25.0\dup{17.5} & 40.0\dup{5.0} & 42.5\dup{30.0} & \textbf{46.8}\dup{22.2} & \textbf{58.2}\dup{27.7} \\
\bottomrule
\end{tabular}
\end{table*}

We test whether routing to the support windows repairs failures that the same intervention placed elsewhere cannot.
After the main comparison (\S\ref{sec:location-frontier}), we move the same budget to other windows (\S\ref{sec:location-matched}) and analyze compositional prompts, repair efficiency, and the link between support overlap and repair success (\S\ref{sec:location-more}).

\subsection{Experimental setup}
\label{sec:location-setup}

\begin{figure}[t]
\centering
\includegraphics[width=\linewidth]{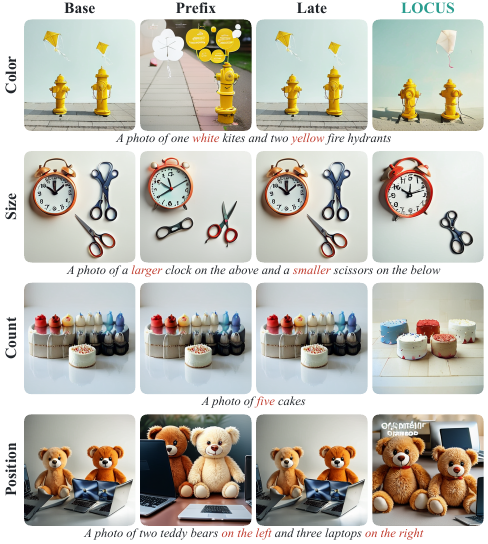}
\caption{\textbf{Qualitative comparison under a matched budget.}
All images are real generations; the target cue is highlighted in red under each row.
Rows cover appearance (color, size) and structure (count, position) failures, and columns compare the base image with the prefix, late, and \method windows.}
\label{fig:failcases}
\end{figure}

\noindent\textbf{Benchmarks and verifiers.}
We evaluate on GenEval~\citep{ghosh2023geneval} and GenEval++~\citep{ye2025echo4o} (280 multi-constraint prompts over seven tags).
GPT-4.1 serves as the outer verifier for diagnosis and scoring, and Qwen3.5-9B~\citep{qwen2026qwen35} re-scores the same images as an independent check (\S\ref{sec:location-matched}).
All arms start from the same base images and differ from them only through the optimized window (details in the appendix).

\noindent\textbf{Window arms.}
To isolate the effect of \emph{location}, we compare five arms (\method, prefix, random, mid, and late windows) under \emph{identical token count, optimizer, reward, and GPT-4.1 verifier}; the only variable is window position.
The prefix arm corresponds to MILR~\citep{mi2025milr}.
The \emph{random} arm draws a window position uniformly per prompt.
The supports of the two regimes together span roughly the first half of the sequence, so a random draw often overlaps one of them; the random arm is therefore a partial-overlap control.

\noindent\textbf{Metrics.}
We report macro accuracy over GenEval++ tags and success@$k$ (S@$k$), the percentage of prompts with a failed base generation that are repaired within $k$ rollouts.
For paired comparisons, we use a stratified paired bootstrap to estimate 95\% confidence intervals and report $p$-values from the exact McNemar test.

\subsection{Main results}
\label{sec:location-frontier}
\label{sec:location-context}

\noindent\textbf{Quantitative comparison.}
Table~\ref{tab:mainresult} compares \method with state-of-the-art generators and test-time methods on GenEval++.
The external systems differ in backbone and training, so the controlled comparison is Janus-Pro versus Janus-Pro + \method.
With no weight update, \method raises the overall accuracy of Janus-Pro from 24.6\% to 46.8\%, with the largest gains in Count (30.0\% to 67.5\%) and Color/Count (12.5\% to 55.0\%).

\noindent\textbf{Qualitative comparison.}
Figure~\ref{fig:failcases} shows two appearance and two structural failures.
In every row only \method repairs the target cue; the prefix and late arms, given the same budget, leave it wrong.

\subsection{Ablation study}
\label{sec:location-matched}

\noindent\textbf{Window location.}
The routing module is the only component that distinguishes \method from prefix optimization, so we ablate it by moving the same budget to other windows.
Table~\ref{tab:matched} reports all five arms on GenEval++.
Routing leads every other arm, and the size of each margin tracks how well the competing window overlaps the attribute's support.
The largest gap is between \method and the late window, at 23.6 percentage points of macro accuracy ($p<0.001$).
\method also scores 6.1 points higher than the random window, with a confidence interval that excludes zero, and 3.2 points higher than the prefix.
Against the late window, 69 prompts are repaired only by \method and 3 only by the late window (18 vs.\ 1 on Count, 17 vs.\ 1 on Color/Count).
Against the prefix, the split is 55 vs.\ 46; the advantage over prefix is largest with a single rollout (\S\ref{sec:location-efficiency}).

\noindent\textbf{Robustness to the verifier.}
Re-scoring the same 1,320 images with Qwen3.5-9B~\citep{qwen2026qwen35} preserves the ordering (Table~\ref{tab:matched}, last column): \method leads the late, mid, random, and prefix windows by 24.5, 20.9, 7.3, and 4.1 points, with McNemar $p<0.05$ for each comparison (appendix).

\begin{table}[t]
\centering
\caption{\textbf{Repair success by rollout budget (\%).}
S@$k$ is the share of failed generations repaired within $k$ rollouts; the optimizer, reward, verifier, and step count are fixed, and the appendix reports all six budgets. Small numbers give the change from MILR.}
\label{tab:successk}
\setlength{\tabcolsep}{4pt}
\begin{tabularx}{\linewidth}{clYYY}
\toprule
{\bf Benchmark} & {\bf Method} & S@1$\uparrow$ & S@3$\uparrow$ & S@6$\uparrow$ \\
\midrule
\multirow{2}{*}{GenEval} & MILR~\cite{mi2025milr} & 37.4\dnone & 61.6\dnone & 74.8\dnone \\
& \cellcolor{groupbg}\textbf{\method} & \cellcolor{groupbg}\textbf{48.1}\dup{10.7} & \cellcolor{groupbg}\textbf{64.9}\dup{3.3} & \cellcolor{groupbg}\textbf{77.1}\dup{2.3} \\
\midrule
\multirow{2}{*}{GenEval++} & MILR~\cite{mi2025milr} & 11.0\dnone & 22.0\dnone & 27.5\dnone \\
& \cellcolor{groupbg}\textbf{\method} & \cellcolor{groupbg}\textbf{14.5}\dup{3.5} & \cellcolor{groupbg}\textbf{25.9}\dup{3.9} & \cellcolor{groupbg}\textbf{32.3}\dup{4.8} \\
\bottomrule
\end{tabularx}
\end{table}

\subsection{Analysis}
\label{sec:location-more}

\noindent\textbf{Compositional prompts.}
\label{sec:location-composition}
Table~\ref{tab:matched} separates the two single-attribute GenEval++ tags (Color, Count) from the five multi-attribute tags.
Combining attributes halves the base accuracy, from 37.5\% to 19.5\%, and \method more than doubles accuracy on the multi-attribute tags, to 40.5\%, with the largest gain on Color/Count (12.5\% to 55.0\%).
The profiles predict where routing should matter most.
On the three multi-attribute tags that pair a structural with an appearance attribute (Color/Count, Color/Pos, Pos/Size), the failed attribute may fall in either regime, so no fixed window covers both.
There \method averages 45.0\%, against 38.3\% for the prefix, 35.0\% for the random window, 27.5\% for the mid window, and 21.7\% for the late window.

\noindent\textbf{Repair efficiency.}
\label{sec:location-efficiency}
\method also improves low-budget repair success and reduces the number of rounds needed to reach a fix (Table~\ref{tab:successk}).
On GenEval, a single rollout repairs 48.1\% of the failed prompts with \method and 37.4\% with prefix optimization.
The mean number of rounds to a fix drops from 2.53 to 2.17, a reduction of 14\%.
Across all GenEval prompts, success after one rollout is 87.9\% with \method and 85.3\% with prefix optimization (paired bootstrap 95\% CI of the difference, 0.1--5.2 points).
On GenEval++, \method leads at all six budgets (appendix).

\begin{figure}[t]
\centering
\includegraphics[width=\linewidth]{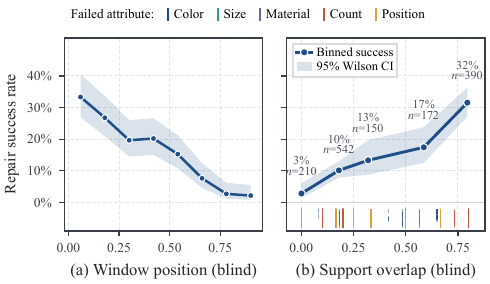}
\caption{\textbf{Support overlap predicts repair success} in the blind sweep (1,464 prompt--window pairs).
(a)~Success by window position. (b)~Success by support overlap; marks show individual scores, colored by failed attribute. Bands are 95\% Wilson CIs.}
\label{fig:overlapsweep}
\end{figure}

\noindent\textbf{Support overlap predicts repair success.}
\label{sec:location-overlap}
If the profiles are causal, repair success should rise with how much of the failed attribute's support a window covers, however the window is chosen.
We score the overlap of a patched step set $W$ with the failed attribute's profile as $\mathrm{overlap}=\tfrac{1}{|W|}\sum_{t\in W} C_a(t)$, using the sliding-window $C_a$ of \S\ref{sec:profile}.
To keep the profile out of window placement, we run a blind sweep.
For each of the 183 failed prompts with measured profiles, we patch eight equal-length windows centered at fixed fractions $\{0.06,0.18,\dots,0.90\}$ of the raster, with the same base images, sampling protocol, and verifier, giving 1,464 prompt--window pairs.
Repair success falls from 33\% in the earliest window to 2\% in the latest (Figure~\ref{fig:overlapsweep}a), the early-commit ordering the profiles predict, recovered without any profile input.
Across overlap bins, success rises from 3\% to 32\% (Figure~\ref{fig:overlapsweep}b; Spearman 0.26, point-biserial 0.27, $p<0.001$ for each with 20,000 permutations).
The five matched arms show the same gradient: over 915 prompt--arm pairs, success rises from 7\% to 36\% across overlap quartiles (both correlations 0.31; appendix), and the random arm, which only partly overlaps the support, falls between \method and the late window accordingly.

\raggedbottom
\section{Conclusion}
\label{sec:conclusion}

Compositional prompts are where autoregressive image generators fail most, and we asked where along the generation each of their attributes can still be changed.
Cross-prompt patching gives each attribute a \profile over the image-token sequence.
Count and position are fixed within the first quarter of the sequence, whereas color, size, and material are decided later, as the object body is decoded, and this support follows where the object sits in the token grid rather than raster order.
This placement accounts for the readability--steerability gap and for why prefix optimization repairs some failures and misses others.
\method uses the measured profiles to decide which latents to optimize at test time.
With the generator frozen, it raises Janus-Pro's GenEval++ accuracy from 24.6\% to 46.8\%, and under the same budget it outperforms windows placed elsewhere in the sequence; even when windows are placed without the profiles, repair success rises with support overlap.
The profiles are currently measured offline for a fixed set of attributes; estimating them per prompt is left to future work.

\section*{LLM Usage Statement}

We used Claude (Anthropic), accessed through the Cursor editor, and ChatGPT (OpenAI), only to polish the writing and to revise the figures and tables.
The authors designed the method, ran all experiments, produced all numbers, and are responsible for every scientific claim and citation.

{
    \small
    \bibliographystyle{ieeenat_fullname}
    \bibliography{main}

@article{chen2025janus,
  title={Janus-Pro: Unified Multimodal Understanding and Generation with Data and Model Scaling},
  author={Chen, Xiaokang and Wu, Zhiyu and Liu, Xingchao and Pan, Zizheng and Liu, Wen and Xie, Zhenda and Yu, Xingkai and Ruan, Chong},
  journal={arXiv preprint arXiv:2501.17811},
  year={2025}
}

@article{sun2024autoregressive,
  title={Autoregressive Model Beats Diffusion: Llama for Scalable Image Generation},
  author={Sun, Peize and Jiang, Yi and Chen, Shoufa and Zhang, Shilong and Peng, Bingyue and Luo, Ping and Yuan, Zehuan},
  journal={arXiv preprint arXiv:2406.06525},
  year={2024}
}

@article{wang2024emu3,
  title={Emu3: Next-Token Prediction is All You Need},
  author={Wang, Xinlong and Zhang, Xiaosong and Luo, Zhengxiong and Sun, Quan and Cui, Yufeng and Wang, Jinsheng and Zhang, Fan and Wang, Yueze and Li, Zhen and Yu, Qiying and others},
  journal={arXiv preprint arXiv:2409.18869},
  year={2024}
}

@article{chameleon2024,
  title={Chameleon: Mixed-Modal Early-Fusion Foundation Models},
  author={{Chameleon Team}},
  journal={arXiv preprint arXiv:2405.09818},
  year={2024}
}

@article{liu2024luminamgpt,
  title={Lumina-{mGPT}: Illuminate Flexible Photorealistic Text-to-Image Generation with Multimodal Generative Pretraining},
  author={Liu, Dongyang and Zhao, Shitian and Zhuo, Le and Lin, Weifeng and Xin, Yi and Li, Xinyue and Qin, Qi and Qiao, Yu and Li, Hongsheng and Gao, Peng},
  journal={arXiv preprint arXiv:2408.02657},
  year={2024}
}

@article{tian2024var,
  title={Visual Autoregressive Modeling: Scalable Image Generation via Next-Scale Prediction},
  author={Tian, Keyu and Jiang, Yi and Yuan, Zehuan and Peng, Bingyue and Wang, Liwei},
  journal={Advances in Neural Information Processing Systems},
  volume={37},
  year={2024}
}

@inproceedings{xie2024showo,
  title={Show-o: One Single Transformer to Unify Multimodal Understanding and Generation},
  author={Xie, Jinheng and Mao, Weijia and Bai, Zechen and Zhang, David Junhao and Wang, Weihao and Lin, Kevin Qinghong and Gu, Yuchao and Chen, Zhijie and Yang, Zhenheng and Shou, Mike Zheng},
  booktitle={International Conference on Learning Representations (ICLR)},
  year={2025}
}

@article{xie2025showo2,
  title={Show-o2: Improved Native Unified Multimodal Models},
  author={Xie, Jinheng and Yang, Zhenheng and Shou, Mike Zheng},
  journal={Advances in Neural Information Processing Systems},
  volume={38},
  year={2025}
}

@article{deng2025bagel,
  title={Emerging Properties in Unified Multimodal Pretraining},
  author={Deng, Chaorui and Zhu, Deyao and Li, Kunchang and Gou, Chenhui and Li, Feng and Wang, Zeyu and Zhong, Shu and Yu, Weihao and Nie, Xiaonan and Song, Ziang and Shi, Guang and Fan, Haoqi},
  journal={arXiv preprint arXiv:2505.14683},
  year={2025}
}

@inproceedings{wu2025harmon,
  title={Harmonizing Visual Representations for Unified Multimodal Understanding and Generation},
  author={Wu, Size and Zhang, Wenwei and Xu, Lumin and Jin, Sheng and Wu, Zhonghua and Tao, Qingyi and Liu, Wentao and Li, Wei and Loy, Chen Change},
  booktitle={Proceedings of the IEEE/CVF International Conference on Computer Vision (ICCV)},
  year={2025}
}

@article{wu2025qwenimage,
  title={Qwen-Image Technical Report},
  author={Wu, Chenfei and Li, Jiahao and Zhou, Jingren and Lin, Junyang and others},
  journal={arXiv preprint arXiv:2508.02324},
  year={2025}
}

@misc{qwen2026qwen35,
  author={{Qwen Team}},
  title={{Qwen3.5}: Towards Native Multimodal Agents},
  howpublished={\url{https://qwen.ai/blog?id=qwen3.5}},
  month={February},
  year={2026}
}

@article{park2025gridar,
  title={Progress by Pieces: Test-Time Scaling for Autoregressive Image Generation},
  author={Park, Joonhyung and Jang, Hyeongwon and Kim, Joowon and Yang, Eunho},
  journal={arXiv preprint arXiv:2511.21185},
  year={2025}
}

@article{vandenoord2017vqvae,
  title={Neural Discrete Representation Learning},
  author={van den Oord, Aaron and Vinyals, Oriol and Kavukcuoglu, Koray},
  journal={Advances in Neural Information Processing Systems},
  volume={30},
  year={2017}
}

@inproceedings{esser2021vqgan,
  title={Taming Transformers for High-Resolution Image Synthesis},
  author={Esser, Patrick and Rombach, Robin and Ommer, Bj{\"o}rn},
  booktitle={Proceedings of the IEEE/CVF Conference on Computer Vision and Pattern Recognition (CVPR)},
  pages={12873--12883},
  year={2021}
}

@article{zheng2022movq,
  title={{MoVQ}: Modulating Quantized Vectors for High-Fidelity Image Generation},
  author={Zheng, Chuanxia and Vuong, Long Tung and Cai, Jianfei and Phung, Dinh},
  journal={Advances in Neural Information Processing Systems},
  volume={35},
  pages={23412--23425},
  year={2022}
}

@article{ho2022cfg,
  title={Classifier-Free Diffusion Guidance},
  author={Ho, Jonathan and Salimans, Tim},
  journal={arXiv preprint arXiv:2207.12598},
  year={2022}
}

@inproceedings{radford2021clip,
  title={Learning Transferable Visual Models From Natural Language Supervision},
  author={Radford, Alec and Kim, Jong Wook and Hallacy, Chris and Ramesh, Aditya and Goh, Gabriel and Agarwal, Sandhini and Sastry, Girish and Askell, Amanda and Mishkin, Pamela and Clark, Jack and Krueger, Gretchen and Sutskever, Ilya},
  booktitle={International Conference on Machine Learning (ICML)},
  pages={8748--8763},
  year={2021}
}

@article{mi2025milr,
  title={MILR: Improving Multimodal Image Generation via Test-Time Latent Reasoning},
  author={Mi, Yapeng and Zhao, Yanpeng and Li, Hengli and Li, Chenxi and Wu, Huimin and Ma, Xiaojian and Zhu, Song-Chun and Wu, Ying Nian and Li, Qing},
  journal={arXiv preprint arXiv:2509.22761},
  year={2025}
}

@article{li2025seek,
  title={Seek in the Dark: Reasoning via Test-Time Instance-Level Policy Gradient in Latent Space},
  author={Li, Hengli and Li, Chenxi and Wu, Tong and Zhu, Xuekai and Wang, Yuxuan and Yu, Zhaoxin and Jiang, Eric Hanchen and Zhu, Song-Chun and Jia, Zixia and Wu, Ying Nian and others},
  journal={arXiv preprint arXiv:2505.13308},
  year={2025}
}

@inproceedings{guo2025twig,
  title={Thinking-while-Generating: Interleaving Textual Reasoning throughout Visual Generation},
  author={Guo, Ziyu and Zhang, Renrui and Li, Hongyu and Zhang, Manyuan and Chen, Xinyan and Wang, Sifan and Feng, Yan and Pei, Peng and Heng, Pheng-Ann},
  booktitle={Proceedings of the IEEE/CVF Conference on Computer Vision and Pattern Recognition (CVPR)},
  pages={26295--26305},
  year={2026}
}

@article{chen2026latentmorph,
  title={Show, Don't Tell: Morphing Latent Reasoning into Image Generation},
  author={Chen, Harold Haodong and Yin, Xinxiang and Shu, Wen-Jie and Zhang, Hongfei and Zhang, Zixin and Liao, Chenfei and Guo, Litao and Chen, Qifeng and Chen, Ying-Cong},
  journal={arXiv preprint arXiv:2602.02227},
  year={2026}
}

@article{zhai2026lac,
  title={Latent Action Control for Reasoning-Guided Unified Image Generation},
  author={Zhai, Fuxiang and Chen, Sixiang and Li, Yingjin and Li, Shuaibo and Lai, Jianyu and Huang, Tengjun and Zhu, Lei},
  journal={arXiv preprint arXiv:2605.16961},
  year={2026}
}

@inproceedings{chen2025scalingar,
  title={{ScalingAR}: Scaling Confidence for Autoregressive Image Generation},
  author={Chen, Harold Haodong and Wu, Xianfeng and Shu, Wen-Jie and Guo, Rongjin and Lan, Disen and Yang, Harry and Chen, Ying-Cong},
  booktitle={International Conference on Machine Learning (ICML)},
  year={2026}
}

@inproceedings{fang2025got,
  title={{GoT}: Unleashing Reasoning Capability of {MLLM} for Visual Generation and Editing},
  author={Fang, Rongyao and Duan, Chengqi and Wang, Kun and Huang, Linjiang and Li, Hao and Tian, Hao and Yan, Shilin and Yu, Weihao and Zeng, Xingyu and Dai, Jifeng and Liu, Xihui and Li, Hongsheng},
  booktitle={Advances in Neural Information Processing Systems (NeurIPS)},
  year={2025}
}

@inproceedings{meral2024conform,
  title={{CONFORM}: Contrast is All You Need for High-Fidelity Text-to-Image Diffusion Models},
  author={Meral, Tuna Han Salih and Simsar, Enis and Tombari, Federico and Yanardag, Pinar},
  booktitle={Proceedings of the IEEE/CVF Conference on Computer Vision and Pattern Recognition (CVPR)},
  pages={9005--9014},
  year={2024}
}

@article{huang2025t2icompbenchpp,
  title={{T2I-CompBench++}: An Enhanced and Comprehensive Benchmark for Compositional Text-to-Image Generation},
  author={Huang, Kaiyi and Duan, Chengqi and Sun, Kaiyue and Xie, Enze and Li, Zhenguo and Liu, Xihui},
  journal={IEEE Transactions on Pattern Analysis and Machine Intelligence},
  volume={47},
  number={5},
  pages={3563--3579},
  year={2025}
}

@inproceedings{binyamin2025count,
  title={Make It Count: Text-to-Image Generation with an Accurate Number of Objects},
  author={Binyamin, Lital and Tewel, Yoad and Segev, Hilit and Hirsch, Eran and Rassin, Royi and Chechik, Gal},
  booktitle={Proceedings of the IEEE/CVF Conference on Computer Vision and Pattern Recognition (CVPR)},
  pages={13242--13251},
  year={2025}
}

@inproceedings{huang2026prove,
  title={Perturb, Repair, Verify: Self-Play Vision-Language Verifiers for Compositional Understanding},
  author={Huang, Zijun and Du, Yingjun and Snoek, Cees G. M.},
  booktitle={Advances in Neural Information Processing Systems (NeurIPS)},
  year={2026}
}

@article{meng2022locating,
  title={Locating and Editing Factual Associations in {GPT}},
  author={Meng, Kevin and Bau, David and Andonian, Alex and Belinkov, Yonatan},
  journal={Advances in Neural Information Processing Systems},
  volume={35},
  pages={17359--17372},
  year={2022}
}

@inproceedings{zhang2023activation,
  title={Towards Best Practices of Activation Patching in Language Models: Metrics and Methods},
  author={Zhang, Fred and Nanda, Neel},
  booktitle={International Conference on Learning Representations (ICLR)},
  year={2024}
}

@inproceedings{cunningham2023sparse,
  title={Sparse Autoencoders Find Highly Interpretable Features in Language Models},
  author={Cunningham, Hoagy and Ewart, Aidan and Riggs, Logan and Huben, Robert and Sharkey, Lee},
  booktitle={International Conference on Learning Representations (ICLR)},
  year={2024}
}

@article{bricken2023monosemanticity,
  title={Towards Monosemanticity: Decomposing Language Models With Dictionary Learning},
  author={Bricken, Trenton and Templeton, Adly and Batson, Joshua and Chen, Brian and Jermyn, Adam and Conerly, Tom and Turner, Nicholas and Anil, Cem and Denison, Carson and Askell, Amanda and others},
  journal={Transformer Circuits Thread},
  year={2023}
}

@article{vig2020causal,
  title={Investigating Gender Bias in Language Models Using Causal Mediation Analysis},
  author={Vig, Jesse and Gehrmann, Sebastian and Belinkov, Yonatan and Qian, Sharon and Nevo, Daniel and Singer, Yaron and Shieber, Stuart},
  journal={Advances in Neural Information Processing Systems},
  volume={33},
  pages={12388--12401},
  year={2020}
}

@inproceedings{wang2022ioi,
  title={Interpretability in the Wild: a Circuit for Indirect Object Identification in {GPT-2} Small},
  author={Wang, Kevin and Variengien, Alexandre and Conmy, Arthur and Shlegeris, Buck and Steinhardt, Jacob},
  booktitle={International Conference on Learning Representations (ICLR)},
  year={2023}
}

@article{alain2016probes,
  title={Understanding Intermediate Layers Using Linear Classifier Probes},
  author={Alain, Guillaume and Bengio, Yoshua},
  journal={arXiv preprint arXiv:1610.01644},
  year={2016}
}

@article{elazar2021amnesic,
  title={Amnesic Probing: Behavioral Explanation with Amnesic Counterfactuals},
  author={Elazar, Yanai and Ravfogel, Shauli and Jacovi, Alon and Goldberg, Yoav},
  journal={Transactions of the Association for Computational Linguistics},
  volume={9},
  pages={160--175},
  year={2021}
}

@inproceedings{hewitt2019control,
  title={Designing and Interpreting Probes with Control Tasks},
  author={Hewitt, John and Liang, Percy},
  booktitle={Proceedings of the 2019 Conference on Empirical Methods in Natural Language Processing (EMNLP)},
  pages={2733--2743},
  year={2019}
}

@article{belinkov2022probing,
  title={Probing Classifiers: Promises, Shortcomings, and Advances},
  author={Belinkov, Yonatan},
  journal={Computational Linguistics},
  volume={48},
  number={1},
  pages={207--219},
  year={2022}
}

@article{li2023inference,
  title={Inference-Time Intervention: Eliciting Truthful Answers from a Language Model},
  author={Li, Kenneth and Patel, Oam and Vi{\'e}gas, Fernanda and Pfister, Hanspeter and Wattenberg, Martin},
  journal={Advances in Neural Information Processing Systems},
  volume={36},
  pages={41451--41530},
  year={2023}
}

@article{turner2023actadd,
  title={Steering Language Models With Activation Engineering},
  author={Turner, Alexander Matt and Thiergart, Lisa and Leech, Gavin and Udell, David and Vazquez, Juan J. and Mini, Ulisse and MacDiarmid, Monte},
  journal={arXiv preprint arXiv:2308.10248},
  year={2023}
}

@article{zou2023repe,
  title={Representation Engineering: A Top-Down Approach to AI Transparency},
  author={Zou, Andy and Phan, Long and Chen, Sarah and Campbell, James and Guo, Phillip and Ren, Richard and Pan, Alexander and Yin, Xuwang and Mazeika, Mantas and Dombrowski, Ann-Kathrin and others},
  journal={arXiv preprint arXiv:2310.01405},
  year={2023}
}

@inproceedings{rimsky2024caa,
  title={Steering {Llama} 2 via Contrastive Activation Addition},
  author={Rimsky, Nina and Gabrieli, Nick and Schulz, Julian and Tong, Meg and Hubinger, Evan and Turner, Alexander Matt},
  booktitle={Proceedings of the 62nd Annual Meeting of the Association for Computational Linguistics (ACL)},
  year={2024}
}

@inproceedings{meng2021sdedit,
  title={{SDEdit}: Guided Image Synthesis and Editing with Stochastic Differential Equations},
  author={Meng, Chenlin and He, Yutong and Song, Yang and Song, Jiaming and Wu, Jiajun and Zhu, Jun-Yan and Ermon, Stefano},
  booktitle={International Conference on Learning Representations (ICLR)},
  year={2022}
}

@inproceedings{hertz2022prompt,
  title={Prompt-to-Prompt Image Editing with Cross Attention Control},
  author={Hertz, Amir and Mokady, Ron and Tenenbaum, Jay and Aberman, Kfir and Pritch, Yael and Cohen-Or, Daniel},
  booktitle={International Conference on Learning Representations (ICLR)},
  year={2023}
}

@article{chefer2023attendandexcite,
  title={Attend-and-Excite: Attention-Based Semantic Guidance for Text-to-Image Diffusion Models},
  author={Chefer, Hila and Alaluf, Yuval and Vinker, Yael and Wolf, Lior and Cohen-Or, Daniel},
  journal={ACM Transactions on Graphics (TOG)},
  volume={42},
  number={4},
  pages={1--10},
  year={2023},
  publisher={ACM New York, NY, USA}
}

@inproceedings{li2024critical,
  title={Critical Windows: Non-Asymptotic Theory for Feature Emergence in Diffusion Models},
  author={Li, Marvin and Chen, Sitan},
  booktitle={International Conference on Machine Learning (ICML)},
  year={2024}
}

@article{tinaz2025emergence,
  title={Emergence and Evolution of Interpretable Concepts in Diffusion Models},
  author={Tinaz, Berk and Fabian, Zalan and Soltanolkotabi, Mahdi},
  journal={Advances in Neural Information Processing Systems},
  volume={38},
  year={2025}
}

@inproceedings{huang2025tide,
  title={{TIDE}: Temporal-Aware Sparse Autoencoders for Interpretable Diffusion Transformers in Image Generation},
  author={Huang, Victor Shea-Jay and Zhuo, Le and Xin, Yi and Wang, Zhaokai and Wang, Fu-Yun and Wang, Yuchi and Zhang, Renrui and Gao, Peng and Li, Hongsheng},
  booktitle={Proceedings of the AAAI Conference on Artificial Intelligence (AAAI)},
  pages={435--443},
  year={2026}
}

@article{kim2025concept,
  title={Concept Steerers: Leveraging K-Sparse Autoencoders for Test-Time Controllable Generations},
  author={Kim, Dahye and Ghadiyaram, Deepti},
  journal={arXiv preprint arXiv:2501.19066},
  year={2025}
}

@article{ghosh2023geneval,
  title={GenEval: An Object-Focused Framework for Evaluating Text-to-Image Alignment},
  author={Ghosh, Dhruba and Hajishirzi, Hannaneh and Schmidt, Ludwig},
  journal={Advances in Neural Information Processing Systems},
  volume={36},
  pages={52132--52152},
  year={2023}
}

@inproceedings{zhang2025omniverifier,
  title={Generative Universal Verifier as Multimodal Meta-Reasoner},
  author={Zhang, Xinchen and Zhang, Xiaoying and Wu, Youbin and Cao, Yanbin and Zhang, Renrui and Chu, Ruihang and Yang, Ling and Yang, Yujiu and Shi, Guang},
  booktitle={International Conference on Learning Representations (ICLR)},
  year={2026}
}

@article{ye2025echo4o,
  title={Echo-4o: Harnessing the Power of {GPT}-4o Synthetic Images for Improved Image Generation},
  author={Ye, Junyan and Jiang, Dongzhi and Wang, Zihao and Zhu, Leqi and Hu, Zhenghao and Huang, Zilong and He, Jun and Yan, Zhiyuan and Yu, Jinghua and Li, Hongsheng and He, Conghui and Li, Weijia},
  journal={arXiv preprint arXiv:2508.09987},
  year={2025}
}

@inproceedings{jiang2025t2ir1,
  title={{T2I-R1}: Reinforcing Image Generation with Collaborative Semantic-level and Token-level {CoT}},
  author={Jiang, Dongzhi and Guo, Ziyu and Zhang, Renrui and Zong, Zhuofan and Li, Hao and Zhuo, Le and Yan, Shilin and Heng, Pheng-Ann and Li, Hongsheng},
  booktitle={Advances in Neural Information Processing Systems (NeurIPS)},
  year={2025}
}

@inproceedings{zhuo2025reflectionflow,
  title={From Reflection to Perfection: Scaling Inference-Time Optimization for Text-to-Image Diffusion Models via Reflection Tuning},
  author={Zhuo, Le and Zhao, Liangbing and Paul, Sayak and Liao, Yue and Zhang, Renrui and Xin, Yi and Gao, Peng and Elhoseiny, Mohamed and Li, Hongsheng},
  booktitle={Proceedings of the IEEE/CVF International Conference on Computer Vision (ICCV)},
  pages={15329--15339},
  year={2025}
}

@inproceedings{jiang2026genagent,
  title={{GenAgent}: Scaling Text-to-Image Generation via Agentic Multimodal Reasoning},
  author={Jiang, Kaixun and Wang, Yuzheng and Zhou, Junjie and Li, Pandeng and Liu, Zhihang and Xie, Chen-Wei and Chen, Zhaoyu and Zheng, Yun and Zhang, Wenqiang},
  booktitle={European Conference on Computer Vision (ECCV)},
  year={2026}
}

@inproceedings{kim2026fire,
  title={{FiRe}: Fine-grained Multimodal Reasoning for Enhanced Image Generation},
  author={Kim, Yongjin and Oh, Yoonjin and Kim, Yerin and Kim, Hyomin and Yun, Jeeyoung and Heo, Yujung and Kim, Minjun and Kim, Sungwoong},
  booktitle={Advances in Neural Information Processing Systems (NeurIPS)},
  year={2026}
}

@inproceedings{esser2024sd3,
  title={Scaling Rectified Flow Transformers for High-Resolution Image Synthesis},
  author={Esser, Patrick and Kulal, Sumith and Blattmann, Andreas and Entezari, Rahim and M{\"u}ller, Jonas and Saini, Harry and Levi, Yam and Lorenz, Dominik and Sauer, Axel and Boesel, Frederic and others},
  booktitle={International Conference on Machine Learning},
  year={2024}
}

@misc{flux2024,
  author={{Black Forest Labs}},
  title={{FLUX}},
  howpublished={\url{https://github.com/black-forest-labs/flux}},
  year={2024}
}

@misc{openai2025gpt4oimage,
  author={{OpenAI}},
  title={Introducing 4o Image Generation},
  howpublished={\url{https://openai.com/index/introducing-4o-image-generation/}},
  year={2025}
}
}

\maketitlesupplementary
\appendix
\section{Appendix}
\label{sec:appendix}

This appendix collects extended protocol details, the stress tests of the causal support profile, and the full LlamaGen-XL commit curves that would otherwise interrupt the main thread.

\subsection{Probe and sparse-autoencoder details (main paper \S3.2)}
This subsection reports the full probe and sparse-autoencoder numbers behind the readability column of Table~1 and the features shown in Figure~3 of the main paper.
The 10-way linear probe predicts the rendered color from image-token latents with 67.1\% accuracy, compared with 10\% by chance.
The cross-space probe direction (colored minus colorless initial latents) yields a margin gap of about 9.
The undercomplete sparse autoencoder has 256 units and a reconstruction FVU of 0.037. A probe on the SAE codes reaches 66.3\% accuracy.
The autoencoder learns 148 monosemantic features that together cover all ten colors. Each feature fires on at least 10\% of examples.

\subsection{Analysis of the causal support profile}
\label{sec:profile-analysis}

The repair results rely on the profiles of \S3 in the main paper.
We now test whether those profiles could be explained away: by a generic perturbation effect, by the coarse scalar \texttt{cross05}, by the raster decode order, or by the calibration set, evaluator, and model used to measure them.

\subsubsection{Negative controls}
\label{sec:profile-controls}

The coverage effect of main-paper \S3.2 is specific to both the attribute and its support window.
Those flips could in principle reflect a generic perturbation, in which any donor activation applied broadly nudges the decoder.
Two negative controls, run on the same six clean color pairs at a single layer ($L{=}0$), rule out this account (Table~\ref{tab:negctrl}).
Replacing the color donor with a \emph{size} donor (\textit{small}$\leftrightarrow$\textit{large}) over the identical full window never flips the color (0/6).
Using the correct color donor but restricting it to the prefix window also never flips it (0/6).
The matched color donor over the full trajectory, in contrast, flips every pair (6/6).
A donor state therefore overrides an attribute only when it both carries the matching attribute and is applied over that attribute's support window.
This joint requirement confirms that the effect tracks attribute-specific causal support rather than arbitrary intervention.

\begin{table}[t]
\centering
\caption{\textbf{Negative controls for color patching.}
The experiment uses six clean minimal pairs and patches at layer 0. The matched arm is the positive control.
A flip requires \emph{both} the matched attribute (color, not size) and the matched support window (full trajectory, not prefix).}
\label{tab:negctrl}
\begin{tabular*}{\linewidth}{@{\extracolsep{\fill}}llcc}
\toprule
{\bf Arm} &
{\bf Donor} &
{\bf Window} &
{\bf Pairs flipped} \\
\midrule
Matched & color & full trajectory & 6/6 \\
Wrong attribute & size & full trajectory & 0/6 \\
Wrong window & color & 11-token prefix & 0/6 \\
\bottomrule
\end{tabular*}
\end{table}

The window requirement is not specific to color, and it holds at larger sample size.
We repeat the specificity controls on 22--24 clean minimal pairs per attribute for color, size, and material (Table~\ref{tab:negctrl-multi}).
Full-trajectory patching with a matched donor flips 77--96\% of pairs, depending on the attribute. With prefix-only patching, the highest flip rate is 14\%.
The wrong-attribute donor flips 5 of 24 color pairs and 4 of 23 material pairs.
For size, it flips 10 of 22 pairs. We attribute this higher rate to the noisy CLIP \textit{small}/\textit{large} readout, which reacts to broader changes in appearance.
Across the 69 pairs, matching the donor and using the full trajectory gives the highest flip rate for each attribute.

\begin{table}[t]
\centering
\caption{\textbf{Support-window requirement beyond color.}
All patches are at layer 0. The matched condition uses the attribute's own donor over the full trajectory. The wrong-attribute condition changes the donor, and the wrong-window condition restricts the matched donor to the prefix.
Parentheses give the number of flips.}
\label{tab:negctrl-multi}
\setlength{\tabcolsep}{4pt}
\begin{tabular*}{\linewidth}{@{\extracolsep{\fill}}lcccc}
\toprule
\multirow{2}{*}{\bf Attribute} &
\multicolumn{3}{c}{\bf Flip rate (\%)} &
\multirow{2}{*}{\bf $n$} \\
\cmidrule(lr){2-4}
& {\bf Matched} & {\bf Wrong attr.} & {\bf Wrong window} & \\
\midrule
color    & 96 (23) & 21 (5)  & 8 (2)  & 24 \\
size     & 77 (17) & 45 (10) & 14 (3) & 22 \\
material & 91 (21) & 17 (4)  & 0 (0)  & 23 \\
\bottomrule
\end{tabular*}
\end{table}

\subsubsection{A commit-time spectrum across 13 attributes}
\label{sec:profile-spectrum}

Broadening the measurement of main-paper \S3.2 to 13 attributes places them on a single early$\to$late spectrum (Table~\ref{tab:cross05}).
Ordered by \texttt{cross05} (main paper \S3.1), structural attributes sit earliest, object-level appearance in the middle, and surface/pose details (\textit{action}, \textit{expression}, \textit{wetness}) latest.
This ordering matches the autoregressive progression ``lay out, then fill appearance, then finish surface.''
Late-committing attributes have near-zero prefix override, so we expect prefix optimization to struggle with them.

\begin{table}[t]
\centering
\caption{\textbf{Commit-time spectrum over 13 attributes.}
We linearly interpolate between grid points to estimate where the cumulative-prefix flip rate first reaches 50\%. Brackets give the 95\% confidence intervals from 2,000 bootstrap resamples of the minimal pairs. The column $n$ gives the number of clean pairs.
Eval. denotes the evaluator: detector (det.) or CLIP image--text model.}
\label{tab:cross05}
\setlength{\tabcolsep}{5pt}
\begin{tabular*}{\linewidth}{@{\extracolsep{\fill}}llcl}
\toprule
{\bf Attribute} &
{\bf Class} &
{\bf \shortstack{\texttt{cross05}\\{}[95\% CI]}} &
{\bf $n$ / Eval.} \\
\midrule
count           & structure       & 48 [24, 352]   & 8 / det. \\
position        & structure       & 84 [72, 168]   & 3 / det. \\
pattern         & surface texture & 154 [96, 213]     & 14 / CLIP \\
age             & appearance      & 160 [132, 216]    & 10 / CLIP \\
size            & appearance      & 163 [128, 327]    & 12 / CLIP \\
color           & appearance      & 186 [156, 220]    & 11 / CLIP \\
object/presence & structure       & 216 [130, 456]    & 11 / det. \\
expression      & facial detail   & 228 [180, 264]    & 16 / CLIP \\
material        & appearance      & 232 [184, 402]    & 9 / CLIP \\
condition       & surface         & 240 [151, 352]    & 14 / CLIP \\
cleanliness     & surface         & 250 [192, 312]    & 12 / CLIP \\
wetness         & surface state   & 256 [168, 336]    & 14 / CLIP \\
action          & pose/relation   & 384 [312, 439]    & 10 / CLIP \\
\bottomrule
\end{tabular*}
\end{table}

On all 13 curves, the flip rate stays at or above 50\% after the first crossing. Count and position reach the threshold first, and action reaches it last.
The bootstrap intervals are wide and overlap through the middle of the spectrum, so the point estimates do not rank color, size, and material against one another.
Count's interval $[24, 352]$ is the widest.
Its curve sits at 50\% from step 48 through step 96, and resampling its eight pairs moves the crossing across that flat stretch.
This is why the causal claims rest on the sliding-window and 2D analyses (\S\ref{sec:raster}) rather than on a single scalar.
Figure~5(a) of the main paper plots these same fine-grid curves for \textit{count}, \textit{position}, \textit{size}, \textit{color}, and \textit{material}; panel (b) is the fixed 144-step windows.

\subsubsection{Support is spatial and semantic, not a decode-time artifact}
\label{sec:raster}

The profiles in main-paper \S3.2 are measured along the decode-step axis, which invites a deflationary objection.
Because Janus-Pro emits tokens in raster order (row $=\lfloor t/24\rfloor$), ``color commits at step $K$'' could merely restate ``the object's pixels decode at step $K$.''
Under that reading, the profile would be a trivial consequence of token ordering rather than an attribute-level causal structure.
We refute this objection directly.

To disentangle decode time from image space, we place the same object at three vertical positions (top / center / bottom) and resolve the causal support over (object placement $\times$ image row-band).
We measure both \emph{sufficiency} (patch a band $\to$ does the color flip?) and \emph{necessity} (leave a band out $\to$ does the flip rate drop?), averaged over eight objects (main paper, Figure~7).
A pure raster/decode-time artifact makes a sharp prediction: the causal support should sit on the \emph{last}-decoded tokens, namely the bottom rows.
The measured support does not.

Table~\ref{tab:raster} gives the corresponding quantitative readout over 19 object-placement conditions.
Patching a single four-row band flips color in 37\% of pairs at rows 4--8 and 32\% at rows 8--12. Patching rows 16--24 yields no flips.
Necessity, the drop in flip rate when a band is left out of a full-coverage patch, is small on every band because coverage is redundant: any single band is usually dispensable.
Leaving out rows 12--16 causes the largest drop, at 13 percentage points.
The necessity peak also shifts downward as the object moves top$\to$center, indicating partial spatial tracking rather than a fixed decode-time band.

\begin{table}[t]
\centering
\caption{\textbf{Per-band 2D causal support map.}
We test eight objects across 19 conditions, with six top, seven center, and six bottom placements.
Row bands are listed in raster decode order, from the first to the last decoded rows.
Sufficiency is the flip rate when patching only that four-row band. Necessity is the full-coverage flip rate minus the rate after leaving out that band, in percentage points (pp).
Each entry pools the 38 pairs and equals the sample-size-weighted mean of the three columns in Figure~7 of the main paper.
Sufficiency peaks in the upper-middle rows and is zero on the last-decoded bottom rows, the opposite of a raster artifact.}
\label{tab:raster}
\begin{tabular*}{\linewidth}{@{\extracolsep{\fill}}lcc}
\toprule
{\bf Row band} &
{\bf Sufficiency (\%)} &
{\bf Necessity (pp)} \\
\midrule
rows 0--4 (first)    & 5  & 5 \\
rows 4--8            & 37 & 11 \\
rows 8--12           & 32 & 11 \\
rows 12--16          & 16 & 13 \\
rows 16--20          & 0  & $-3$ \\
rows 20--24 (last)   & 0  & 0 \\
\bottomrule
\end{tabular*}
\end{table}

Thus, what we called a ``commit time'' is more precisely a spatial/semantic \emph{support region} of the token grid.
This finding reframes the profile and removes the most natural confound.
The structure we measure is attribute- and content-driven, not a restatement of raster decode order.

\subsubsection{Profiles transfer across objects, evaluators, and models}
\label{sec:profile-transfer}

A profile measured in a single setting could still reflect calibration fit, evaluator choice, or model specificity.
We therefore test it under three controls: held-out objects, a unified evaluator, and two further AR models.
These controls complement the second-verifier check on the repair results (\S\ref{sec:location-verifier}).

We first test held-out transfer.
If the support window were only the place we tuned, it should not transfer to unseen instances.
We calibrate the color window on one set of minimal pairs and apply it, count-matched, to a held-out pool of 40 disjoint objects paired with held-out colors.
After excluding two objects with an unclean base render, we score $n{=}38$ (Table~\ref{tab:heldout}).
The predicted window flips the held-out color in 34.2\% of cases. The late window patches the same number of steps and yields no flips ($p=0.0002$).
The predicted window versus the other valid windows is a non-significant trend, mirroring the matched-window ordering of main-paper \S4.3.
This transfer indicates that the calibrated window behaves as a transferable causal topology rather than a post-hoc lookup.

\begin{table}[t]
\centering
\caption{\textbf{Held-out transfer to disjoint objects.}
We score 38 of 40 sampled objects and patch 96 tokens in each arm.
The calibrated predicted window flips held-out colors while a late window flips none; \method vs.\ other valid windows is a non-significant trend.}
\label{tab:heldout}
\begin{tabular*}{\linewidth}{@{\extracolsep{\fill}}lcc}
\toprule
{\bf Arm} &
{\bf Flip rate (\%)} &
{\bf 95\% CI (\%)} \\
\midrule
\method (ours) & 34.2 & [18.4, 50.0] \\
early (prefix) & 26.3 & [13.2, 39.5] \\
random & 21.1 & [7.9, 34.2] \\
late & 0.0 & [0.0, 0.0] \\
\bottomrule
\end{tabular*}
\end{table}

We next test evaluator dependence.
The commit-time profiles read appearance with the CLIP image--text model and structure with GenEval detectors.
The structure-early/appearance-late ordering could therefore in principle be an artifact of using two different judges.
We rule out this artifact by re-judging the \emph{same} patched images with a single unified vision-language model (VLM; OmniVerifier-7B).
For every clean (pair, window) patch, we ask the VLM one targeted question about the attribute value and recompute the override curve and its \texttt{cross05} (Table~\ref{tab:vlmsanity}).
The unified judge reproduces the original ordering: the structural attribute commits earliest (\textit{count} \texttt{cross05}$=120$ under both judges), while the appearance attributes commit far later (\textit{color} 216, \textit{material} 208).
The ``structure early, appearance late'' ordering is thus a property of the model rather than of the evaluator split.

\begin{table}[t]
\centering
\caption{\textbf{Unified-evaluator sanity check on the commit profile.}
We compute \texttt{cross05} with the original readouts (CLIP for appearance and a detector for structure) and with OmniVerifier-7B, which scores the same patched images. The column $n$ gives the number of clean pairs.
The structure-before-appearance ordering holds under both judges.}
\label{tab:vlmsanity}
\begin{tabular*}{\linewidth}{@{\extracolsep{\fill}}llccc}
\toprule
\multirow{2}{*}{\bf Attribute} &
\multirow{2}{*}{\bf Family} &
\multicolumn{2}{c}{\bf \texttt{cross05}} &
\multirow{2}{*}{\bf $n$} \\
\cmidrule(lr){3-4}
& & {\bf Original} & {\bf VLM} & \\
\midrule
count    & structure  & 120 & 120 & 4 \\
size     & appearance & 134 & 128 & 12 \\
material & appearance & 240 & 208 & 10 \\
color    & appearance & 202 & 216 & 12 \\
\bottomrule
\end{tabular*}
\end{table}

We then test whether the picture is specific to Janus-Pro.
On LlamaGen-XL, a second AR model with T5-prefix conditioning and $T{=}256$ image tokens, we repeat the cumulative-prefix cross-prompt activation patching of main-paper \S3.2.
The test covers one structural and three appearance attributes, with at least 34 clean minimal pairs each (Table~\ref{tab:llamagen}).
Donor and receiver captions have the same number of T5 tokens, and appearance pairs are run in both directions.
A pair is clean only if the unpatched donor and receiver renders read out their own values; patching the whole sequence reproduces the donor image exactly for every item.
Because sequence lengths differ across models, we compare \texttt{cross05} as a percentage of the sequence length.

\begin{table}[t]
\centering
\caption{\textbf{Cross-model commit profile on LlamaGen-XL.}
The sequence contains 256 tokens. Brackets give the 95\% confidence interval from 2,000 bootstrap resamples of the minimal pairs. Parentheses give \texttt{cross05} as a percentage of the sequence. Ovr@16 is the flip rate after patching the first 16 image tokens.
Position is the structural attribute; the other three are appearance attributes.
Position is read out by CLIP on the left/right image halves and the appearance attributes by the CLIP image--text model; the Janus-Pro column gives the corresponding \texttt{cross05} (\% of the sequence) from Table~\ref{tab:cross05}.
Position commits first, and its interval does not overlap color's or material's.}
\label{tab:llamagen}
\setlength{\tabcolsep}{2.5pt}
\begin{tabular*}{\linewidth}{@{\extracolsep{\fill}}lcccc}
\toprule
{\bf Attribute} &
{\bf $n$} &
{\bf \texttt{cross05}} &
{\bf Ovr@16 (\%)} &
{\bf Janus-Pro} \\
\midrule
position & 34 & 16 [16, 32] (6\%)   & 50 & 15--24 \\
size     & 52 & 48 [16, 128] (19\%)  & 42 & 38 \\
color    & 40 & 64 [48, 96] (25\%)   & 7  & 32--38 \\
material & 42 & 128 [80, 128] (50\%) & 10 & 47 \\
\bottomrule
\end{tabular*}
\end{table}

\Needspace{6\baselineskip}
Position reaches a 50\% flip rate first, at 16 tokens (6\% of the sequence). Its confidence interval, 16--32 tokens, lies below those for color and material.
The two regimes of main-paper \S3.2 reappear, with one difference in degree.
As on Janus-Pro, color and material respond weakly to a 16-token prefix, with flip rates of 7\% and 10\%, respectively.
Unlike on Janus-Pro, position is largely set by the prefix: 16 patched tokens flip half the pairs.
A finer grid places the crossing at 12 tokens (95\% CI, 4--24 tokens). A four-token prefix flips 35\% of pairs.
On LlamaGen, the structural layout is thus committed at the very start of the sequence, which is the regime where prefix optimization is expected to work.
Size is the one attribute that does not fit cleanly.
Its CLIP \textit{small}/\textit{large} readout registers flips in 42\% of pairs after 16 steps. The confidence interval for the crossing spans 16--128 tokens.
Under the content-gated override, size moves to $\texttt{cross05}{=}128$, so we do not rank it against the other appearance attributes.

These curves patch cumulative prefixes $[0,K)$, so they establish the commit \emph{order} on LlamaGen but not where within the sequence its support concentrates.
Every tested window starts at the first token.
The band-resolved and two-dimensional analyses of main-paper \S3.2 and \S\ref{sec:raster} are measured on Janus-Pro only.

\begin{figure}[t]
\centering
\includegraphics[width=\linewidth]{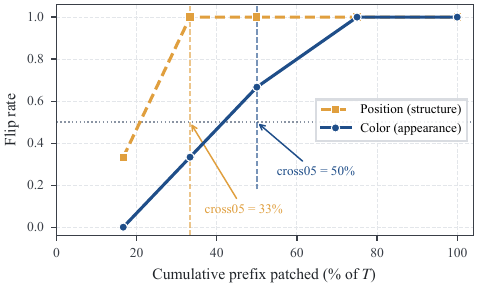}
\caption{\textbf{Lumina-mGPT 2.0.}
Position reaches a 50\% flip rate after patching 33\% of the sequence, compared with 50\% for color. The sequence has 9,216 tokens, and we test 18 pairs per attribute.}
\label{fig:lumina_crossmodel}
\end{figure}

Finally, we probe a \emph{third} architecture, Lumina-mGPT 2.0, and again recover the within-model dichotomy: structure and appearance are measured on the same model with the same intervention.
Lumina-mGPT 2.0~\citep{liu2024luminamgpt} is a Chameleon-style~\citep{chameleon2024} decoder-only transformer whose discrete image tokens live inside the LLM vocabulary and are decoded by a MoVQGAN detokenizer~\citep{zheng2022movq}.
It thus adds a third, independent conditioning/architecture regime alongside Janus-Pro (text-as-tokens with a learned VQ) and LlamaGen (T5-prefix conditioning).
We measure the commit time of \emph{color} (appearance) and \emph{position} (structure) on $n{=}18$ pairs per attribute with the same cumulative-prefix \emph{cross-prompt activation patching} used for Janus-Pro.
Donor activations are transplanted into a receiver run over a growing prefix $[0,K)$ of the $T{=}9216$ image tokens, and we record the override (flip-to-donor) rate versus $K$ (Figure~\ref{fig:lumina_crossmodel}).
Structure commits before appearance.
Position reaches a 50\% flip rate at 3,072 tokens (33\% of the sequence) and saturates there. Color reaches 50\% at 4,608 tokens (half the sequence) and saturates at 75\% coverage.
At every prefix length, the flip rate for position is at least as high as that for color. The swap margins are about 0.10--0.15 for position and 0.03--0.06 for color.

A methodological observation makes this comparison meaningful.
When we instead steer with a single global direction, adding $\alpha\lVert h\rVert\hat{d}$ to every step as in a linear ITI-style edit, appearance is still overridable.
The override appears only \emph{late}, once the injection covers essentially the whole sequence.
Structure, in contrast, is not overridable at all: a global push toward ``swap left/right'' or ``add an object'' does not rearrange the layout or change the count; it simply collapses the image.
A uniform direction cannot perform a position-dependent rearrangement.
Structure is therefore \emph{rigid} to a global edit and yields only under a faithful transplant of the donor's actual early activations.
We consequently report the color-vs-position \texttt{cross05} comparison under the \emph{same} method (patching) for both attributes, so the ordering is not an artifact of mixing interventions.
Across three architecturally distinct AR generators, the same ordering holds: structure is written into the sequence earlier than appearance.
On LlamaGen and Lumina-mGPT 2.0, this evidence covers the commit order only, not where the support concentrates.

\subsection{LlamaGen-XL commit profiles (\S\ref{sec:profile-transfer})}
\label{sec:appendix-llamagen}
Table~\ref{tab:llamagen-full} reports every LlamaGen-XL patching curve, extending Table~\ref{tab:llamagen} with the two attributes left out there.
Patching overwrites the conditional stream's post-norm hidden state (the input to the output head) with the donor's over image tokens $[0,K)$, $K\in\{16,32,48,64,80,96,128,160,192,256\}$; count and two-object presence add $K\in\{8,24,224\}$.
The donor, the unpatched receiver, and all patched windows are rows of one generation, so the base image is the same for every window.
Count uses three receiver$\to$donor pairs (two$\to$three, two$\to$four, three$\to$four) over 75 COCO plural nouns in both directions.
Two-object presence asks whether patching adds a second object that the receiver caption omits.
Both read out with the GenEval Mask2Former detector.
Clean rates are low for position and count because LlamaGen-XL at 256~px rarely renders the requested arrangement or exact count on both donor and receiver.
The content-gated override counts a flip only if the object CLIP score stays within $0.90\times$ of the base render (CLIP readouts), or the frame is not degenerate (detector readouts).
A finer position grid with four-token increments gives a \texttt{cross05} of 12 tokens on the same 34 pairs (95\% CI, 4--24 tokens).

\begin{table*}[!t]
\centering
\caption{\textbf{LlamaGen-XL commit curves for all attributes.}
All curves use cumulative-prefix patching on 256-token sequences. Brackets give 95\% confidence intervals from 2,000 bootstrap resamples of the minimal pairs.
Clean gives the number of pairs passing the clean gate out of the total tested. Ovr@16 is the flip rate after patching the first 16 tokens.
Position commits earliest, while count does not separate from color and two-object presence commits late.}
\label{tab:llamagen-full}
\begin{tabular*}{\textwidth}{@{\extracolsep{\fill}}llcccc}
\toprule
{\bf Attribute} &
{\bf Readout} &
{\bf Clean} &
{\bf \texttt{cross05}, flip} &
{\bf \texttt{cross05}, gated} &
{\bf Ovr@16 (\%)} \\
\midrule
position  & CLIP halves & 34/450 & 16 [16, 32]    & 32 [16, 80]    & 50 \\
size      & CLIP        & 52/112 & 48 [16, 128]   & 128 [48, 256]  & 42 \\
color     & CLIP        & 40/48  & 64 [48, 96]    & 80 [48, 96]    & 7  \\
count     & detector    & 20/450 & 80 [24, 128]   & 80 [24, 128]   & 20 \\
material  & CLIP        & 42/128 & 128 [80, 128]  & 96 [80, 128]   & 10 \\
presence  & detector    & 64/360 & 192 [160, 192] & 192 [160, 192] & 3  \\
\bottomrule
\end{tabular*}
\end{table*}

\begin{table*}[t]
\centering
\caption{\textbf{Repair success by rollout budget (\%).}
We use the same protocol as in Table~4 of the main paper and report all six budgets.}
\label{tab:successk-full}
\begin{tabularx}{\textwidth}{clYYYYYY}
\toprule
{\bf Benchmark} & {\bf Method} & S@1$\uparrow$ & S@2$\uparrow$ & S@3$\uparrow$ & S@4$\uparrow$ & S@5$\uparrow$ & S@6$\uparrow$ \\
\midrule
\multirow{2}{*}{GenEval} & MILR~\cite{mi2025milr} & 37.4 & 56.5 & 61.6 & 67.9 & 71.0 & 74.8 \\
& \cellcolor{groupbg}\textbf{\method} & \cellcolor{groupbg}\textbf{48.1} & \cellcolor{groupbg}\textbf{58.8} & \cellcolor{groupbg}\textbf{64.9} & \cellcolor{groupbg}\textbf{71.0} & \cellcolor{groupbg}\textbf{74.0} & \cellcolor{groupbg}\textbf{77.1} \\
\midrule
\multirow{2}{*}{GenEval++} & MILR~\cite{mi2025milr} & 11.0 & 16.1 & 22.0 & 24.8 & 25.2 & 27.5 \\
& \cellcolor{groupbg}\textbf{\method} & \cellcolor{groupbg}\textbf{14.5} & \cellcolor{groupbg}\textbf{20.0} & \cellcolor{groupbg}\textbf{25.9} & \cellcolor{groupbg}\textbf{29.5} & \cellcolor{groupbg}\textbf{30.0} & \cellcolor{groupbg}\textbf{32.3} \\
\bottomrule
\end{tabularx}
\end{table*}

The confidence intervals for count and color overlap. For count, the flip rate falls from 50\% with 80 patched tokens to 40\% with 96.
Two-object presence commits late (75\% of the sequence), as object presence also sits mid-spectrum on Janus-Pro (Table~\ref{tab:cross05}), so we do not use it as evidence for either regime.

\subsection{Full repair-efficiency curve (main paper \S4.4)}
Table~\ref{tab:successk-full} extends Table~4 of the main paper to every budget from $k{=}1$ to 6.
\method leads MILR at every budget on both benchmarks.

\subsection{Support overlap of the matched arms (main paper \S4.4)}
\begin{figure}[t]
\centering
\includegraphics[width=\linewidth]{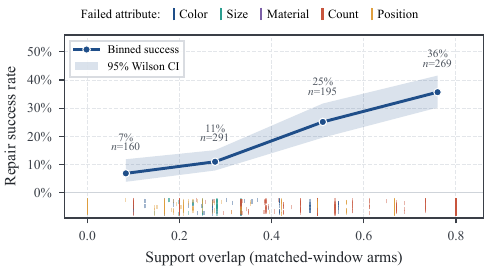}
\caption{\textbf{Support overlap of the matched arms} (915 prompt--arm pairs).
Marks show individual overlap scores, colored by failed attribute. Bands are 95\% Wilson intervals.}
\label{fig:overlapcorr}
\end{figure}

We reconstruct the patched steps $W$ of each of the five matched arms on the 183 failed prompts with measured support profiles, giving 915 prompt--arm pairs, and pair each overlap score with the arm's binary repair outcome (Figure~\ref{fig:overlapcorr}).
The Spearman and point-biserial correlations with repair success are both 0.31 ($p<0.001$ for each, using 20,000 permutations), and success rises from 7\% to 36\% across overlap quartiles.
Because \method reads $C_a$ when it places its window, this correlation is partly built in by design; the blind sweep of main-paper \S4.4 removes that dependence.

\subsection{Second-verifier consistency (main paper \S4.3)}
\label{sec:location-verifier}
The matched-window conclusion was established with GPT-4.1, so we re-score the same 1,320 saved images with an architecturally disjoint verifier, Qwen3.5-9B~\citep{qwen2026qwen35}, at zero additional generation cost (main paper, Table~3).
The ordering survives the swap: routing leads every other window under both judges.
With Qwen3.5, routing leads the late and middle windows by 24.5 and 20.9 percentage points, respectively. Both comparisons have McNemar $p<0.001$ under either verifier.
Against the random window, routing leads by 6.1 percentage points with GPT-4.1 ($p=0.021$) and 7.3 points with Qwen3.5 ($p=0.009$). Both confidence intervals exclude zero.
Routing also leads the prefix by 3.2 points with GPT-4.1 and 4.1 points with Qwen3.5. The latter comparison has $p=0.033$.
The location effect is therefore not an artifact of the GPT-4.1 judge.

\subsection{\method implementation details (main paper \S3.3)}
This subsection reports the optimizer, sampling, and reward settings shared by every repair arm in main-paper \S4.
We optimize the selected image-token hidden states with REINFORCE, using the reward minus the target baseline as the advantage. We use Adam with a learning rate of 0.03 and a classifier-free guidance weight of 5.0.
Each repair round draws 16 candidate rollouts.
Base generations and rollouts use multinomial sampling for all 576 image tokens at a temperature of 1.0, with a fixed seed per prompt.
A rollout therefore differs from its base generation only through the optimized window, where the token is the argmax of the optimized latent.
This design makes every arm a paired comparison against the same base image.
Each round then runs up to 6 inner optimization steps with early stopping, including the main results and the matched-window controls of main-paper \S4.3.
The inner reward is a GPT-4.1 per-dimension partial-credit score; the outer verifier is GPT-4.1 (re-scored by Qwen3.5-9B in \S\ref{sec:location-verifier}).
All runs use a single NVIDIA A100 GPU.
Optimizer and sampling settings follow MILR~\citep{mi2025milr}, and the prefix arm corresponds to MILR.

\subsection{A capability hierarchy of attribute control}
\label{sec:hierarchy}
\begin{figure}[t]
\centering
\includegraphics[width=\linewidth]{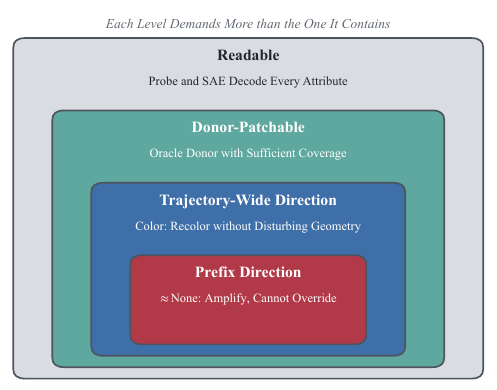}
\caption{\textbf{Capability hierarchy of attribute control in AR image generation.}
Each level is strictly harder than the one it contains: attributes are \emph{readable} everywhere, all are \emph{donor-patchable} given an oracle donor and sufficient coverage, only color is controllable by a single \emph{trajectory-wide} fixed direction, and \emph{prefix} fixed-direction control is essentially empty.
The remaining attributes require support-matched, reward-guided optimization.}
\label{fig:hierarchy}
\end{figure}

Our findings organize attribute control into nested levels, each demanding more than the last (Figure~\ref{fig:hierarchy}): readable $\supsetneq$ donor-patchable $\supsetneq$ trajectory-wide fixed-direction controllable $\supsetneq$ prefix fixed-direction controllable.
The outer levels are readable (probe/SAE) and donor-patchable (all attributes, given an oracle donor and sufficient sequence coverage).
Trajectory-wide fixed-direction control holds for color, which is local, redundant, and re-decidable token-by-token, so a single global direction applied across the trajectory can recolor without disturbing geometry.
Prefix fixed-direction steering mainly amplifies attributes that the prompt already requests (main paper \S3.2).
Size, material, and count demand coordinated, position-dependent structural change that no single global direction can produce, which is why they fall out of the fixed-direction level entirely.
This separation tracks the information-flow account of main-paper \S3.3: a global geometric constraint is fixed early and cannot be locally rewritten, while a locally re-read appearance stays malleable across the trajectory.
This account explains why representation-engineering steering succeeds only narrowly and why support-matched, reward-guided optimization is needed for the remaining attributes.

\subsection{Limitations}
\label{sec:limitations}
These limitations bound the scope of the claims.
First, several mechanism experiments use a single layer ($L{=}0$) and modest sample sizes for some attributes.
Second, appearance and structure use different evaluators (CLIP vs.\ detectors).
We mitigate this evaluator mismatch with within-class window profiles, a per-claim evaluator label, and a unified-VLM sanity check that reproduces the structure-early/appearance-late ordering (Table~\ref{tab:vlmsanity}), but do not fully remove it.
Third, the cross-model evidence covers commit order only: on LlamaGen-XL and Lumina-mGPT 2.0 we measure cumulative-prefix curves, not the band-resolved or two-dimensional support.
Fourth, content gating uses a fixed threshold; we have not characterized sensitivity across thresholds such as 0.90, 0.92, and 0.95.
The necessary condition is scoped to test-time \emph{image-token latent} interventions under a fixed model, reward, and budget, not to weight-level edits or arbitrary budgets.

\end{document}